\documentclass{elsarticle}

\usepackage{hyperref}

\journal{Neural Networks}

\usepackage{amsmath,amssymb,amsthm}
\usepackage{bm}

\usepackage{dsbda} 

\usepackage{algorithm}
\usepackage{algpseudocode}

\usepackage{booktabs}
\usepackage{rotating}
\usepackage{multirow}

\usepackage{xcolor}
\hypersetup{
    colorlinks,
    linkcolor={red!50!black},
    citecolor={blue!50!black},
    urlcolor={blue!80!black}
}

\newcommand{\revise}[2]{#2} 

\newtheorem{definition}{Definition}

\newcommand{\node}[0]{vertex\xspace}
\newcommand{\nodes}[0]{vertices\xspace}

\newcommand{\Nodes}[0]{Vertices\xspace}

\newcommand{\dataset}[1]{#1}

\newcommand{\wos}{\dataset{PharmaBio}\xspace}
\newcommand{\dblpeasy}{\dataset{DBLP-easy}\xspace}
\newcommand{\dblphard}{\dataset{DBLP-hard}\xspace}

\newcommand{\Dt}[1]{\mathrm{tdiff}_{#1}}
\newcommand{\degree}{\mathrm{deg}}
\newcommand{\neighbors}{\mathcal{N}}
\DeclareMathOperator{\tsmin}{\operatorname{time}}

\newcommand{\vs}{vs.\xspace}
\newcommand{\embrace}[1]{{(#1)}}

\newcommand\myTitle{Lifelong Learning on Evolving Graphs Under the Constraints of Imbalanced Classes and New Classes}

\begin{document}
\begin{frontmatter}

\title{\myTitle{}\tnoteref{publnote}}
\tnotetext[publnote]{Accepted manuscript (after peer review, before copy-editing). Published article available at \url{https://doi.org/10.1016/j.neunet.2023.04.022}}

\author[mpi]{Lukas Galke\fnref{zbwnote}\corref{mycorrespondingauthor}}
\cortext[mycorrespondingauthor]{Corresponding author}
\ead{lukas.galke@mpi.nl}

\author[umc]{Iacopo Vagliano}
\ead{i.vagliano@amsterdamumc.nl}

\author[uulm]{Benedikt Franke}
\ead{benedikt.franke@uni-ulm.de}

\author[uulm]{Tobias Zielke}
\ead{tobias-1.zielke@uni-ulm.de}

\author[uulm]{Marcel Hoffmann}
\ead{marcel.hoffmann@uni-ulm.de}

\author[uulm]{Ansgar Scherp}
\ead{ansgar.scherp@uni-ulm.de}

\address[mpi]{Max Planck Institute for Psycholinguistics, Nijmegen, Netherlands}

\address[umc]{Amsterdam UMC, location University of Amsterdam, Netherlands}

\address[uulm]{University of Ulm, Germany}

\fntext[zbwnote]{Parts of this research were carried out while L.G. was with ZBW -- Leibniz Information Centre for Economics, Kiel, Germany}

\begin{abstract}
Lifelong graph learning deals with the problem of continually adapting graph neural network (GNN) models to changes in evolving graphs.  We address two critical challenges of lifelong graph learning in this work: dealing with new classes and tackling imbalanced class distributions.  The combination of these two challenges is particularly relevant since newly emerging classes typically resemble only a tiny fraction of the data, adding to the already skewed class distribution.  We make several contributions: First, we show that the amount of unlabeled data does not influence the results, which is an essential prerequisite for lifelong learning on a sequence of tasks.  Second, we experiment with different label rates and show that our methods can perform well with only a tiny fraction of annotated nodes.  Third, we propose the gDOC method to detect new classes under the constraint of having an imbalanced class distribution. The critical ingredient is a weighted binary cross-entropy loss function to account for the class imbalance. Moreover, we demonstrate combinations of gDOC with various base GNN models such as GraphSAGE, Simplified Graph Convolution, and Graph Attention Networks. Lastly, our k-neighborhood time difference measure provably normalizes the temporal changes across different graph datasets.  With extensive experimentation, we find that the proposed gDOC method is consistently better than a naive adaption of DOC to graphs. \revise{}{Specifically, in experiments using the smallest history size, the out-of-distribution detection score of gDOC is 0.09 compared to 0.01 for DOC. Furthermore, gDOC achieves an Open-F1 score, a combined measure of in-distribution classification and out-of-distribution detection, of 0.33 compared to 0.25 of DOC (32\% increase).}
\end{abstract}


\end{frontmatter}

\section{Introduction}\label{sec:introduction}
Graph representation learning has gained momentum in recent years~\cite{Hamilton2020}.
Significant developments have been made on graph neural networks (GNNs) based on the seminal work by Scarselli ~et\,al.~\cite{DBLP:journals/tnn/ScarselliGTHM09} in 2009. In particular, the work on graph convolution~\cite{DBLP:journals/corr/KipfW16,DBLP:conf/nips/HamiltonYL17} and graph attention~\cite{velickovic2018graph} triggered a wave of works that turned GNNs from a niche topic into one of the most active research fields in machine learning~\cite{Hamilton2020}.

The enormous interest in graph representation learning is motivated by the flexibility of graphs to represent virtually any kind of real-world data and the ability to model relationships between data points, \ie vertices, rather than just the properties of independent and identically distributed (i.\,i.\,d.) data points.

A common challenge in machine learning, and thus in graph representation learning, for tasks such as vertex classification is an imbalance in the class distribution.
For example, the popular Cora citation dataset~\cite{DBLP:journals/aim/SenNBGGE08} with seven classes has a heavily skewed class distribution.
The smallest class makes about $7 \%$ of the vertices, while the largest constitutes about $30 \%$.
Citation graphs grow over time.
While new publications and citations appear over time, new classes in the form of new scientific fields emerge.
In numerous cases, real-world graph data evolves, with new classes, vertices, and edges appearing over time.
Generally, this requires the machine learning model to deal with changes and continually adapt the model to new tasks.
Adapting a model to new tasks is investigated under the term of lifelong machine learning~\cite{thrun1998lifelong,lifelonglearningbook}. 
Unsurprisingly, lifelong learning on graph data is also recently gaining more and more interest
~\cite{DBLP:journals/nn/ParisiKPKW19,wangStreamingGraphNeural2020,zhouOvercomingCatastrophicForgetting2021,wangLifelongGraphLearning2021,galke2021lifelong,chenTrafficStreamStreamingTraffic2021,febrinantoGraphLifelongLearning2022}. 
Numerous applications can benefit from lifelong graph learning, including social networks, traffic prediction, recommender systems, and anomaly detection~\cite{febrinantoGraphLifelongLearning2022}. 

Existing works on lifelong graph learning ~\cite{wangStreamingGraphNeural2020,zhouOvercomingCatastrophicForgetting2021,wangLifelongGraphLearning2021,galke2021lifelong,chenTrafficStreamStreamingTraffic2021,DBLP:conf/www/Cai0GTXZ022} were mainly concerned with catastrophic forgetting, \ie how to adapt a model to new data without forgetting what it had learned before.
For a recent survey on lifelong graph learning, we refer to~\cite{febrinantoGraphLifelongLearning2022}.
In our prior work, we developed an incremental training procedure to continuously maintain graph representation learning models during the evolution of a graph~\cite{galke2021lifelong}. 
However, the existing body of works, including our own, does not yet address \textit{detection of new classes in the lifelong graph learning scenario}, \ie dealing with the adaptation of the graph representation under the emergence of new classes while the graph evolves.
Generally, when a graph evolves, and new classes emerge, these new classes are relatively rare compared to the number of vertices of already-known classes.
Having only a few examples for these new classes further exacerbates the challenge of imbalance in the class distribution because the i.\,i.\,d. assumption does not hold for graphs~\cite{Hamilton2020}, particularly when the model continually adapts to changing data.
Instead, different influential factors define a vertex label induced by vertex features and the edges between the vertices.
Thus, it is exciting to investigate the combination of the two challenges of the imbalanced class distribution and the detection of new classes.

We extend our lifelong training procedure for evolving graphs~\cite{galke2021lifelong} with a new open-world learning module, gDOC, to detect the appearance of new classes in the graph.
In particular, we design our gDOC method for detecting new classes to handle imbalanced classes in graph data.
It extends the class detection method Deep Open Classification (DOC)~\cite{DOC} from textual data to graph data.
We experimentally demonstrate that gDOC can detect new classes in graph data while maintaining high accuracy for classifying in-distribution vertices.
The overall performance of gDOC for out-of-distribution (OOD) detection plus vertex classification is consistently higher than a plain DOC.
The key to the success of gDOC is weighting the binary cross-entropy loss to counter the imbalanced graph data.
Furthermore, we show how to train and retrain graph models to cope with changes, newly emerging classes, and different label rates.
We demonstrate that inductively pre-trained graph models are robust to adding unlabeled data.
This insight is an essential prerequisite for successful lifelong learning on graph data.
For comparability of different temporal datasets, we introduced the $k$-neighborhood time differences measure~\cite{galke2021lifelong}.
The measure enables a selection of history sizes in lifelong graph learning that accounts for of the dataset's temporal granularity. We prove that our measure fulfills this critical equivariance property.

\subsection{Problem Formalization: Lifelong Learning on Graphs}
\label{sec:problem-formalization}
The critical question in lifelong learning is whether it is helpful to maintain a single model throughout a sequence of tasks versus retraining a new model from scratch for the next task. 
We call the former case  ``warm restarts'', which means that the initial model parameters for the current task come from the final parameters from the previous task.
This reuse of parameter values from the current task to the next task is called \textit{internal knowledge}.
The latter case is the ``cold restarts'' scenario, in which we train a new model from random initialization for each task.

Lifelong learning, \ie maintaining a single model over time~\cite{DBLP:conf/ijcai/ThrunM95,thrun1998lifelong}, is only beneficial when warm restarts are at least as good as cold starts under comparable training budgets.
In contrast to internal knowledge, there is also \textit{external knowledge}, which is the data used for incremental training.
The amount of external knowledge available, \ie the past graph data, is determined by a \textit{history size}. \revise{}{Note that this history is not separate from the actual data. If the label of a past vertex changes, this change will be reflected in the next incremental training step.}
The history size is, in turn, determined based on the temporal granularity of the considered graph and the time differences within the receptive field of the GNN.
The number of GNN layers defines the receptive field of a GNN. Each layer corresponds to one hop. Thus, the receptive field comprises the $k$-hop neighborhood of each \node.
Combined, the temporal granularity of the graph and the receptive field allow one to provide comparable results across datasets with different evolution speeds. 

We define the problem of new class detection of a graph's vertices in an evolving graph as a form of open-world lifelong graph learning~\cite{lifelonglearningbook,galke2021lifelong}.
We employ this understanding of lifelong graph learning in four different settings. We introduce the four settings below and refer to the corresponding sections for experimental details.
Our notation is summarized in Table~\ref{tab:notation}.

\begin{table}[t]
  \centering
  \caption{Summary of our Notation}
  \label{tab:notation}
  \begin{tabular}{lp{10cm}}
  \toprule
  $\mX$ & A matrix holding the \node features of each \node as rows\\
  $\vy$ & A vector holding the label of each \node as its entries\\
  $\sY_t$ & The set of classes at time $t$   \\
  $\gT$ & A task composed of the graph $\gG$ along with \node features $\mX$ and \node labels $\vy$\\
  $\gT_t$ & Task $t$ within a sequence of tasks\\
  $\gG_t$ & State of the graph at time $t$ with \nodes $V_t$ and edges $E_t$\\
  $c$ & The history size used for training the GNN \\
  $c(V, E, t)$ & A function to determine a history size depending on a set of vertices $V$ and edges $E$ with time information $t$\\
  $\tilde{\gG}_t$ & The trimmed graph with respect to the history size $c$, \ie older \nodes and edges are removed\\
  $\tilde{\mX}$ & A matrix holding \node features, but with rows removed that correspond to \nodes removed in $\tilde{\gG}_t$.\\
  $\tilde{\vy}$ & A vector holding \node labels, but with entries removed that correspond to \nodes removed in $\tilde{\gG}_t$.\\
  \bottomrule
  \end{tabular}
\end{table}

\begin{figure}[ht!]
    \centering
    \includegraphics[width=0.8\linewidth]{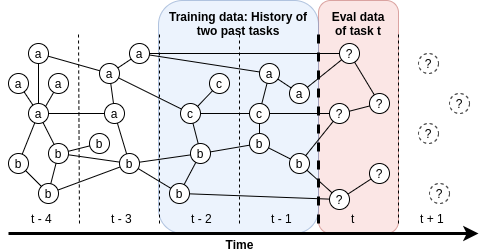}
    \caption{Illustration of the problem of lifelong graph learning~\cite{galke2021lifelong} with class imbalance and new classes.
    At each time $t$, the learner has to classify new vertices of task $\gT_t$ (red). 
    Any task might come with previously unseen classes.
    For example, the class ``$c$'' emerged only at task $t-2$ and was subsequently added to the class set. 
    The learner may use internal and external knowledge from previous tasks to adapt to the current task.
    After evaluating task $\gT_t$, we continue with task $\gT_{t+1}$.
    }\label{fig:teaser}
\end{figure}

\begin{definition}[Lifelong Learning~\cite{lifelonglearningbook}]
A learner has to perform a possibly open-ended sequence of learning tasks $\gT_1, \gT_2, \ldots, \gT_T$. At each time $t$, the learner is faced with a new learning task $\gT_{t}$, for which it may use the (internal and external) knowledge $\gK$ that it has been provided with and accumulated in the previous tasks $\gT_1, \gT_2, \ldots, \gT_{t-1}$. 
\end{definition}

We cast this definition into a lifelong graph learning problem by considering each task $\gT_t := (\gG_t, \mX^\embrace{t}, \vy^\embrace{t})$ to be a \node classification task with graph $\gG_t = (V_t, E_t)$, corresponding \node features $\mX^{(t)} \in \mathbb{R}^{\mid V_t \mid \times D}$, and \node labels $\vy^{(t)} \in \mathbb{N}^{\mid V_t \mid}$.
We denote the set of all classes available at time $t$ as $\sY_t$. 
We assume that the class distribution $\sY_t$ is skewed and that this distribution changes over time, \eg by adding new classes.
Thus, any of the \nodes that appeared with the graph $\gG_t$ may have new, so-far \textit{unseen} classes.
This means that $\sY_t$ may 
contain classes that were not in $\sY_{t-1}$.

To ensure that past knowledge is helpful to perform the task $\gT_t$, we impose $\gG_{t-1} \cap \gG_t \neq \emptyset$. 
This means there is at least some overlap in the two graphs $\gG_{t-1}$ and $\gG_t$ of two consecutive tasks. 
We assume that the vertices' features and labels do not change after they have appeared, \ie it holds $\mX^{(t-1)}_u = \mX^{(t)}_u, \vy^{(t-1)}_u = \vy^{(t)}_u$ if $u \in V_{t-1} \cap V_{t}$. 

In order to control, \ie limit the amount, of explicit knowledge available for a task $\gT_t$, we introduce the history size $c$. 
We set $\tilde{\gT}_t := (\tilde{\gG}_t, \tilde{\mX}^{(t)}, \tilde{\vy}^{(t)})$ with $\tilde{\gG}_t := \gG_t \setminus ( \gG_1 \cup \gG_2 \cup \ldots \cup \gG_{t-c-1})$, \ie we remove vertices (with their features and labels) and edges connecting these vertices 
that are ``older'' than $t-c-1$ to construct $\tilde{\mX}_t$ and $\tilde{\vy}_t$. 
Still, the model may use implicit knowledge acquired by the model parameters through warm restarts in earlier tasks for the task $\tilde{\gT}_t$.

Based on this formalization, we consider four settings for lifelong learning on graphs:
We briefly describe each experimental setting below and refer to the corresponding sections for the details. 

\paragraph{Two-task Setting}
The two-task setting is a simplified setting of lifelong graph learning where we only consider two tasks $\gT_1$ and $\gT_2$ \revise{}{without new classes, \ie $\sY_1 = \sY_2$}. 
This setting is suitable for applying our approach to any (non-temporal) standard dataset by defining $\gT_1$ as the training graph and $\gT_2$ as the test graph.
We use this setup to compare transductive and inductive learning on graphs in \Secref{exp:transductive-inductive}.
\revise{The idea is to compare models that have been inductively pre-trained on the labeled subgraph of the training data against models that have been trained transductively on the training data plus the unlabeled test data.}{The goal is to test the effect of including unlabeled
test data during training, which is relevant for the following settings.}

\paragraph{Task-sequence Setting}
In this setting, one is provided with a sequence of tasks \revise{}{$\tilde\gT_1, \tilde\gT_2, \ldots, \tilde\gT_T$, each with a limited history size}.
We assume that new classes appear over time\revise{}{, \ie $\sY_t$ not necessarily equals $\sY_{t-1}$}.
Although new classes are present in the task $\tilde\gT_t$'s data in this setting, no methods are employed to detect the new classes.
\revise{}{Furthermore, the ground-truth labels $\vy^{(t-1)}$ to $\vy^{(t-1-c)}$ are available, when training for the next task $\tilde\gT_t$.}
This setting is investigated in the experiments of \Secref{exp:lifelong-learning}.

\paragraph{Task-sequence Setting with Limited Labeled Data}
\revise{}{This setting is the same as the previous one with the difference that we relax the assumption that all past labels $\tilde\vy$ are available. Here, only a fraction of labels becomes available when training for the next task $\tilde\gT_t$ instead of the labels of all vertices in the history.}
This setting is reflected in the experiments of \Secref{exp:limited-labeled-data}.

\paragraph{Task-sequence Setting with Unseen Class Detection}
In the final variant, we analyze the capabilities of the GNN models to detect new classes.
\revise{}{In addition to lifelong vertex classification as in previous task-sequence settings}, the models now need to emit a binary decision per \node whether it belongs to a previously known class \revise{}{in $\sY_{t-1}$} (in-distribution), or belongs to a new, unseen class \revise{}{$\sY_{t} \setminus \sY_{t-1}$} (out-of-distribution).
This setting is reflected in the experiments of \Secref{exp:open-learning}.

\subsection{Key Contributions}

We analyze different aspects of lifelong graph learning, detecting new classes and training graph models with skewed class distributions.
This research is based on and significantly extends a training procedure and framework for lifelong learning published in our work~\cite{galke2021lifelong,galke2019graph}.
The main findings of our prior work are that warm restarts in lifelong learning enable one to use fewer training data, \ie using a smaller history size. Only a small amount of historical data is necessary to achieve a performance comparable to retraining, \ie cold start on the entire graph.
The key contributions of this work that expand on our previous findings are summarized below.

\paragraph{Method for Detecting New Classes in Lifelong Graph Learning}
  \revise{}{We extend our previous lifelong graph learning framework~\cite{galke2021lifelong} with a generic module to detect new classes. For this module, we compare a naive adaption of DOC from text to graphs with a proposed extension, gDOC, that takes into account class imbalance.
  We find that gDOC outperforms DOC in all cases.
  Specifically, in the lowest history size setting, gDOC achieves an F1 score of 0.33 compared to 0.25 for DOC (32\% increase), and the OOD detection score rises from 0.01 for DOC to 0.09 for gDOC.}

\paragraph{Influence of the Availability of Labels on the Performance}
\revise{}{We investigate different settings of varying availability of labels in lifelong graph learning:
First, we add unlabeled data to the graph after training on the labeled subgraph.
We show that adding unlabeled data does not further increase the performance~\cite{galke2019graph}.   
In the context of lifelong graph learning, this insight shows that graph models only need to be retrained when new \emph{labeled} becomes available.
Second, we vary the label rate between 10\% and 90\% in a task-sequence lifelong learning setting. 
We observe a trend that parameter reuse (warm restarts) is generally preferred over cold starts and becomes even more relevant with lower label rates.}

\paragraph{The $k$-Neighborhood Time Difference Measure is Equivariant to the Temporal Granularity of Evolving Graphs}
\revise{}{
Our $k$-neighborhood time difference measure $\Dt{k}(\gG)$~\cite{galke2021lifelong} captures the temporal differences between connected vertices in evolving graphs. It can be reused independently from the other methods proposed in this work.
We use the $\Dt{k}(\gG)$ measure to determine history sizes comparable between different temporal graphs with differing change behavior (fast versus slow changes). 
Here, we prove that $\Dt{k}(\gG)$ is equivariant to temporal granularity, such as when having monthly versus yearly time information.}

\subsection{Organization of the Article}
Subsequently, we provide an overview of related work.
The extended incremental training algorithm for lifelong graph learning, as well as our new class detection method gDOC, are described in \Secref{sec:methods}.
In \Secref{sec:deltat}, we describe the $k$-neighborhood time differences measure, which we use to determine comparable history sizes across datasets and provide proof that the measure is invariant to different temporal granularities.
The datasets used in our experiments are described and analyzed in \Secref{sec:datasets}.
In Sections~\ref{exp:transductive-inductive} to \ref{exp:open-learning}, we describe the experimental procedure and report the results of our four experiments. 
We perform experiments for each of the four lifelong graph learning settings introduced in Section~\ref{sec:problem-formalization}.
First, in \Secref{exp:transductive-inductive}, we analyze the difference between transductive \vs inductive learning, \ie the influence of adding unlabeled data, on standard (static) datasets, pre-processed in either many-few or few-many train/test splits.
Second, we analyze the case of continually adding labeled data during a sequence of tasks in \Secref{exp:lifelong-learning}.
Third, in \Secref{exp:limited-labeled-data}, we take the most powerful methods and the hardest dataset of the previous experiment to analyze the influence of different label rates.
Lastly, in the experiments reported in \Secref{exp:open-learning}, we employ our gDOC method to automatically detect new classes while still being able to correctly classify the vertices of known classes.
The results are discussed in \Secref{sec:discussion} before we conclude.

\section{Related Work and Selection of Models for Experiments}\label{sec:rw}
Our work connects with various research areas: graph neural networks, lifelong learning, and out-of-distribution detection.

In \Secref{sub:rw:gnn}, we relate our work to the literature on graph neural networks~\cite{Hamilton2020}. Since our aim is that our incremental training algorithm applies to a wide range of GNN models, we seek to obtain high coverage among different types of GNN models in our experiments and select a representative GNN model for each type.
In \Secref{sub:rw:llg}, we relate our work to the lifelong learning literature~\cite{lifelonglearningbook} concerning general approaches for non-graph data and approaches for graph data. 
In \Secref{sub:rw:ood}, we relate our work to the literature on unseen class detection and approaches to the more general problem of out-of-distribution detection while differentiating between supervised and unsupervised, as well as between crisp and scoring approaches.

\subsection{Graph Neural Networks}\label{sub:rw:gnn}
The success of graph convolution~\cite{DBLP:journals/corr/KipfW16} has caused a resurgence of interest in graph neural networks~\cite{DBLP:journals/tnn/ScarselliGTHM09}. 
In a generic formulation, the hidden representation of \node $i$ in layer $l$ is defined as:
$h_i^{(l+1)} = \sigma \left( \sum_{j \in \mathcal{N}(i)} \frac{1}{c_{ij}} W^{(l)} h_j^{(l)} \right)$,
where $\mathcal{N}(\cdot)$ refers to the set of adjacent \nodes and $\sigma$ is a
nonlinear activation function. The normalization factor $c_{ij}$ depends on the
respective model: the original Graph Convolutional Networks (GCN)~\cite{DBLP:journals/corr/KipfW16} use $c_{ij} = \sqrt{\mid \mathcal{N}(i) \mid} \cdot
\sqrt{ \mid \mathcal{N}(j) \mid}$.

To categorize the vast literature on graph neural networks, we adopt the distinction of Dwivedi~\etal{}~\cite{dwivedi2020benchmarking} between isotropic and anisotropic GNN architectures.
In isotropic GNNs, all edges are treated equally, while in anisotropic GNNs, the weights for the edges are dynamically calculated, \eg based on the features of the involved \nodes.
Similarly, we differentiate between standard GNN approaches versus scalable approaches that rely on either subgraph sampling or decoupling the neighborhood aggregation from the neural network component.
Our goal is to understand how different approaches of GNNs react to situations of evolving graphs and new classes with an imbalanced distribution.

\paragraph{Isotropic Graph Neural Networks}
In addition to graph convolutional networks~\cite{DBLP:journals/corr/KipfW16}, examples of isotropic GNNs are GraphSAGE with mean aggregation~\cite{DBLP:conf/nips/HamiltonYL17}, DiffPool~\cite{DBLP:conf/nips/YingY0RHL18}, and GIN~\cite{DBLP:conf/iclr/XuHLJ19}.
We consider GraphSAGE-Mean~\cite{DBLP:conf/nips/HamiltonYL17} as a representative for isotropic GNNs because its special treatment of the \nodes self-connections has been shown to be beneficial~\cite{dwivedi2020benchmarking}.\label{sub:graphsage-mean}
The representations of self-connections are concatenated with averaged neighbors' representations before multiplying the parameters. 
In GraphSAGE-Mean, the procedure for obtaining representations on layer $l+1$ for \node $i$ is given by the equations:
$\hat{\vh}_i^{l+1} = \vh_i^l \mid \mid \frac{1}{\degree_i} \sum_{j \in \neighbors(i)} \vh_j^l$ and $ \vh_i^{l+1} = \sigma(\mU^l \hat{\vh}_i^{l+1})$.

\paragraph{Anisotropic Graph Neural Networks}
Examples of anisotropic GNNs include graph attention networks~\cite{velickovic2018graph}, GatedGCN~\cite{DBLP:journals/corr/abs-1711-07553}, and MoNet~\cite{DBLP:conf/cvpr/MontiBMRSB17}.
We consider Graph Attention Networks (GATs)~\cite{velickovic2018graph} to be representative of the class of anisotropic GNNs.
In GATs, the representations in layer $l+1$ for \node $i$ are computed as follows:
$\hat{\vh}_i^{l+1} = \alpha_{ii}^l \vh_{i}^l + \sum_{j \in \neighbors(i)} \alpha_{ij}^l \vh_{j}^l$ and $\vh_i^{l+1} = \sigma(\mU^l \hat{\vh}_i^{l+1})$, 
where $\neighbors(i)$ is the set of adjacent \nodes to \node $i$, $U^l$ are learnable parameters, and $\sigma$ is a non-linearity. 
The edge weights $\alpha_{ij}$ are calculated using a self-attention mechanism based on $h_i$ and $h_j$, \ie the softmax of $a(\mU^l \vh_i \mid\mid \mU^l \vh_j)$ on the edges, where $a$ is an MLP and $\cdot \mid\mid \cdot$ is the concatenation operation.

\paragraph{Scalable Graph Neural Networks}
There are further approaches that have been specifically proposed to scale GNNs to large graphs.
These approaches fall into two categories: 
decoupling neighborhood
aggregation from the neural network component \cite{DBLP:SimpleGCN,signSampling,10.1145/3394486.3403296,graphmlp,lightgcn} and subgraph sampling \cite{DBLP:conf/nips/HamiltonYL17,adaptiveSampling,fastgcn,clustergcn,DBLP:conf/iclr/ZengZSKP20}.

In simplified GCN (SGC)~\cite{DBLP:SimpleGCN},
the neighborhood aggregation of GNNs is decoupled from the feature transformation.  
In SGC, any non-linearities are removed, and consecutive weight matrices are collapsed into a single one. In more detail, SGC can be described by equation $\mathbf{\hat{Y}}_\mathrm{SGC} = \text{softmax}(\mathbf{S}^K\mathbf{X}\mathbf{\Theta})$,where $\mathbf{S}$ is the normalized adjacency matrix and $\mathbf{\Theta}$ is the weight matrix.
As such, SGC is a scalable variant of Graph Convolutional Networks ~\cite{DBLP:journals/corr/KipfW16} that admits regular minibatch sampling. 
The hyperparameter $K$ has a similar effect as the number of layers in regular GCNs. Instead of using multiple layers, the $k$-hop neighborhood is computed by $\mathbf{S}^K$, so that $\mathbf{\mS}^K \mX$ can be precomputed.
This makes SGC efficient, while, surprisingly, it does not necessarily harm performance~\cite{DBLP:SimpleGCN}.
\revise{}{LightGCN~\cite{lightgcn} is an approach designed for collaborative filtering that entirely removes the feature transformation and nonlinear activation and only builds upon the neighborhood aggregation of GCNs. Since LightGCN is tailored towards collaborative filtering recommender systems, we opt for SGC in our experiments.}

GraphSAINT~\cite{DBLP:conf/iclr/ZengZSKP20} is a state-of-the-art subgraph sampling technique.
In GraphSAINT, entire subgraphs are sampled for training GNNs.
Subgraph sampling introduces a bias that is counteracted by normalization coefficients for the loss function.
We used the best-performing random-walk sampling for our experiments.
The underlying GNN is exchangeable, but the authors suggest using Jumping Knowledge Networks (JKNets)~\cite{DBLP:JKNetwork}. 
JKNets introduce skip connections, or residual connections, to GNNs:
Each hidden layer has a direct connection to the output layer, in which the representations are aggregated, for example, by concatenation.
\revise{}{FastGCN~\cite{fastgcn} is another sampling-based approach, which proposes importance sampling for the assembly of rooted subtree batches. However, GraphSAINT reports a favorable comparison against FastGCN. Thus, we chose GraphSAINT for our experiments.}

\paragraph{Dynamic Graph Methods}

Different GNN methods have been proposed for dynamic graphs. 
An important distinction here is that, in contrast to our problem statement, these methods focus on dealing with varying \node features and labels over time, \eg a user becomes banned from a social network at a specific time~\cite{rossi2020temporal}. 

This body of work includes dynamic embedding methods~\cite{DBLP:conf/www/NguyenLRAKK18,lee2020dynamic}, autoencoder-based methods~\cite{DBLP:journals/corr/abs-1805-11273,DBLP:journals/kbs/GoyalCC20}, GNNs for graphs with a fixed \node set~\cite{DBLP:conf/icml/TrivediDWS17,DBLP:conf/iconip/SeoDVB18,kumar2018learning,DBLP:conf/iclr/TrivediFBZ19,DBLP:journals/pr/ManessiRM20,DBLP:conf/wsdm/SankarWGZY20,rossi2020temporal}, and inductive GNN methods that can deal with previously unseen \nodes~\cite{DBLP:conf/aaai/ParejaDCMSKKSL20,Xu2020Inductive}.
These methods focus on the case of dynamic outputs. This means that a \node can be in class ``a'' at time t and in class ``b'' at time $t+1$. 
In our case, the \node features and labels remain the same over time, but the graph itself is evolving with new \nodes, edges, and classes appearing over time. 
On the contrary, the related approaches assume a static set of \nodes, which makes them inapplicable to the problem of lifelong learning with unseen class detection that we investigate in this paper.

\paragraph{Selection of Representative Base Models}\label{selection-of-base-models}
For our lifelong learning experiments, we systematically select representative GNN architectures and scalable GNN techniques.
From each of these four categories (anisotropic versus isotropic GNNs, and preprocessing versus sampling), we select one representative for our lifelong graph learning experiments.
We chose GraphSAGE as a representative for the class of isotropic GNNs, GAT as a representative for anisotropic GNNs, SGC as a representative for the scaling-via-decoupling approach, and GraphSAINT for the scaling-via-sampling approach. Moreover, we also include JKNets because it is recommended as a base model for GraphSAINT and because its residual connections have been shown to be beneficial~\cite{dwivedi2020benchmarking}.
Lastly, we also include an MLP as a graph-agnostic baseline in all of our experiments.

\subsection{Lifelong Learning}\label{sub:rw:llg}
We first summarize the general literature on lifelong learning, before describing the related work on lifelong learning on graphs.

\paragraph{Lifelong Learning on Non-Graph Data}
Lifelong learning, or continual learning~\cite{DBLP:conf/nips/Lopez-PazR17}, has been present in machine learning research since the mid 1990s~\cite{DBLP:conf/ijcai/ThrunM95,thrun1998lifelong,DBLP:conf/aaaiss/SilverYL13,liu2017lifelong}.
The goal of lifelong learning is to develop approaches that can adapt existing models to new tasks.
Although similar on a superficial level, it differs from online learning~\cite{DBLP:conf/icml/HerbsterPW05}, in which the focus is on processing a data stream efficiently.
Ruvolo and Eaton~\cite{DBLP:conf/icml/RuvoloE13} introduced a lifelong learning algorithm with convergence guarantees that employs multitask learning so that later tasks can improve earlier tasks.
Fei~\etal{}~\cite{DBLP:conf/kdd/FeiW016} analyzed SVMs in a lifelong learning environment and introduced cumulative learning.
Cumulative learning is related to our approach since we consider that some data are shared among the tasks.
Lopez-Paz \& Ranzato~\cite{DBLP:conf/nips/Lopez-PazR17} introduced a gradient episodic memory framework for the image domain, where examples can be processed independently, and address the catastrophic forgetting problem,
\ie the loss of previously learned information when new information is learned~\cite{DBLP:journals/connection/Robins95}.

Similarly to our work, Wang~\etal{}~\cite{wang2021neuralcomput} decompose lifelong learning into the subproblems of rejecting unknown instances, classifying accepted instances, and reducing the cost of learning. However, the work of Wang~\etal{} is on image data, in which the examples are independent of each other, and thus the challenges of dealing with graph data are not reflected.
Another promising approach to lifelong learning, and in particular class-incremental learning, is iCaRL~\cite{Rebuffi_2017_CVPR}, in which prototype vectors of known classes are stored and classification is carried out by taking the nearest distance to these prototypes.
However, applying this method to graph data is nontrivial because the \nodes are not independent from each other, but connected via edges.
For an overview of lifelong learning in general (not specific to graphs), we refer to a recent textbook~\cite{lifelonglearningbook}.

\paragraph{Lifelong Learning on Evolving Graphs}
We now focus on the related work on lifelong learning \emph{on graphs}.
The challenge of dealing with graph data is a special challenge for lifelong learning approaches.
That is because in graphs the nodes are not independent of each other because they are connected through edges.
Related work on lifelong learning \emph{on graphs} is still rather limited.
We refer to Febrinanto~\etal\cite{febrinantoGraphLifelongLearning2022} for a recent survey that covers five recent works on graph lifelong learning.

The most similar approach to ours is Experience Replay GNN~\cite{zhouOvercomingCatastrophicForgetting2021}, which proposed to overcome catastrophic forgetting~\cite{FRENCH1999128}, \ie the problem of previous knowledge being quickly forgotten when models are adjusted to new tasks.
The Replay GNN~adapts to new tasks with the help of an experience replay buffer. 
The buffer holds a subset of the graph that is determined on the basis of different selection strategies: mean of features, coverage maximization, or influence maximization. This work is conceptually similar to our work. However, we use the time information from the nodes in conjunction with a history size to determine which part of the graph is kept in memory. 

Wang et al.~\cite{wangLifelongGraphLearning2021} proposed a very different strategy to tackle lifelong learning on graphs. 
Their main goal was again to alleviate catastrophic forgetting.
The authors explored a preprocessing step that transforms the \node classification task into a graph classification task, \ie each \node is converted into a feature graph. 
Therefore, \nodes become independent so that they can follow the lifelong learning approach from Lopez{-}Paz and Ranzato~\cite{DBLP:conf/nips/Lopez-PazR17} (see above).

Continual-GNN~\cite{wangStreamingGraphNeural2020} addressed the issue of catastrophic forgetting with a regularization approach. The authors detected new patterns in the data (but not involving any new classes) with an information propagation method. Then they used a combination of experience replay and model regularization to avoid catastrophic forgetting.
The result was that their approach leads to performance comparable to model retraining. In relation to this work, we also compare our lifelong-learned models against models retrained from scratch for each task (cold restart) but additionally consider other conditions such as the history size.

Another recent approach~\cite{DBLP:conf/www/Cai0GTXZ022} uses neural architecture search to find a suitable model architecture for lifelong learning on graphs. In particular, the proposed approach focuses on multimodal inputs, such as features extracted via BERT~\cite{DBLP:conf/naacl/DevlinCLT19} and vision transformers~\cite{vit}, rather than dealing with new classes and how to detect them.

Liu~et\,al.~\cite{liuOvercomingCatastrophicForgetting2021} decompose the lifelong learning problem into incremental training over separate tasks, where the class labels are disjunct between tasks (class-incremental). 
The primary aim is to alleviate catastrophic forgetting. 
In contrast, we focus on forward transfer, \ie understanding if previously acquired knowledge is helpful for future tasks given that there is some overlap between classes in the tasks. 
In addition, we consider the detection of new classes as part of the problem statement.

 \revise{}{Tan~et\,al.~\cite{tanGraphFewshotClassincremental2022} formulate a few-shot class-incremental version of the lifelong learning problem statement and a  hierarchical attention-based graph meta-learning approach. 
The work introduces a regularization objective that aims to avoid overfitting to both, the known classes and new class(es).
Their version of the problem statement assumes that some of the new classes' \nodes are annotated with a label. In contrast, there are no annotations for the new classes in our problem statement. In other words, we seek to detect \nodes that do not belong to any of the known classes, while Tan et al.  assume that some labeled information is present such that the few-shot learning setting applies. Both versions have their merits, yet they differ in their possible use cases: 
Tan~et\,al.~aim to integrate new classes with as few labeled data as possible, while we focus on the problem of automatically detecting new classes.}

So far, none of the related works on lifelong learning in graphs
\revise{have}{} considered the problem of detecting unseen classes and
rejecting the classification of the respective \nodes. Knowing when a model is
likely to make mispredictions is a crucial property for deploying reliable
systems in practice, which motivates us to explore the combination of lifelong
learning on graphs and unseen class detection. Moreover, labeled data is often
not fully available in real-world conditions, which we investigate here because
it has not yet been considered in previous work on lifelong graph learning.

\subsection{Out-of-Distribution and Unseen Class Detection}\label{sub:rw:ood}
Unseen class detection, or open-world learning, is considered a subcategory of lifelong learning~\cite{lifelonglearningbook}. Still, more general methods for out-of-distribution (OOD) detection are also related to the problem of detecting unseen classes.

\paragraph{Unsupervised Out-of-Distribution Detection}
A key challenge is that softmax activation, often used as the final layer of classification, leads to highly confident mispredictions even when the input data are far from the training distribution.
To address this, Liang~\etal{}~\cite{DBLP:conf/iclr/LiangLS18} resorted to temperature scaling, while
Lee~\etal{}~\cite{DBLP:conf/nips/LeeLLS18} proposed using the Mahalanoubis distance.
Macedo~\etal~\cite{macedo2019entropic,macedo2021improving} replaced softmax activation with IsoMax activation based on entropy.
However, all these approaches only produce an 
OOD score and neglect the thresholding problem; \ie they cannot produce crisp decisions for each \node whether it belongs to a new class or not.

\paragraph{Supervised Out-of-Distribution Detection}
Other approaches rely on explicit outlier data that can be used for supervised training of the outlier module~ \cite{DBLP:conf/nips/DhamijaGB18,DBLP:conf/iclr/HendrycksMD19}.
This is difficult to apply here because we do not distinguish between out-of-distribution and in-distribution but between previously seen classes and previously unseen classes. When we had appropriate training data for the unseen classes, we could train directly on them rather than considering them as OOD.
For a detailed discussion of OOD methods, we refer to recent surveys~\cite{ood-survey,ad-survey}.

\paragraph{Crisp and Unsupervised Unseen Class Detection}
We are particularly interested in methods that emit a crisp decision on whether
the classification of an instance (a new vertex) should be rejected. In this
regard, there are several approaches to detect new classes using classic
machine learning
methods~\cite{DBLP:journals/tkde/MasudGKHT11,DBLP:conf/cvpr/BendaleB16,DBLP:conf/kdd/FeiW016}.
For example, Wu~\etal~\cite{DBLP:conf/icdm/WuPZ20} have used variational graph
autoencoders for uncertain \node representation learning. They generate
multiple versions of features and test the certainty of a \node belonging to a
known class.

In Deep Open Classification (DOC)~\cite{DOC}, the authors proposed a method for
the detection of new classes in text categorization. To perform the detection,
the final softmax activation of a neural network is replaced by elementwise
sigmoid activation. Then, they derived a threshold for unseen class detection
by measuring the logits' standard deviation across the training set. Their
experiments on datasets with balanced classes indicated that DOC is preferable
to OpenMax~\cite{DBLP:conf/cvpr/BendaleB16} and
cbsSVM~\cite{DBLP:conf/kdd/FeiW016}. 

\revise{}{Xu~et\,al.~\cite{l2ac} propose L2AC  for open-world learning in product classification with text data.
The L2AC framework is composed of a ranker and a meta-classifier. The ranker retrieves examples from seen classes, which are fed into the meta-classifier to classify the current example or reject its classification.
The meta-classifier consists of a matching layer and an aggregation layer.
The one-vs-many matching layer determines the similarity to each known class via the top-$k$ known examples. The aggregation layer, a many-to-one BiLSTM, merges the $k$ similarity values  into an OOD score per class. After a final fully-connected layer, the classification rule is similar to the one of Deep Open Classification: reject if the score of all classes falls below a threshold of $0.5$, or else assign the class label with the maximum logit.
Thus, the class detection of L2AC corresponds to a special case of both DOC and gDOC with a fixed threshold of $0.5$ and without risk reduction.
}

\revise{}{
Reusing existing OOD detection methods for graphs with interconnected nodes (non-i.i.d.) and imbalanced class distributions is not straightforward. While technically possible, combining graph neural networks with standard OOD methods is rarely evaluated. Here, we transfer the most promising method that is capable of crisp new class detection, DOC, from text to graphs, along with an extension to account for class imbalance.}

\subsection{Summary}
\revise{}{To summarize, lifelong learning on graphs is a new research topic with only 
a few previous works. The previous works on lifelong graph learning mainly tackle catastrophic forgetting in class- or data-incremental settings on standard datasets. In contrast, we focus on forward transfer, \ie whether and how much previous knowledge is helpful for future tasks, and use evolving graph datasets close to assumed applications
None of the discussed works analyzes the problem of new class detection on graph data we tackle in this work. Moreover, we analyze the effects of label rate, history size, and parameter reuse in incremental learning on evolving graphs with selected representative GNN base models to obtain a complete picture.
Although OOD methods are related to new class detection, dealing with interconnected vertices in graphs with imbalanced class distributions is a new challenge, which we tackle in this work.}

\section{Lifelong and Open-World Graph Learning}\label{sec:methods}
We explore the combination of lifelong learning on evolving graphs and new class detection on evolving graphs with imbalanced class distributions.
In the following, we recapitulate the incremental training algorithm~\cite{galke2021lifelong}, which we extend by a generic module for unseen class detection.
Then, we introduce our gDOC method, an extension of DOC~\cite{DOC} for unseen class detection, and show how it can be used in conjunction with incrementally trained graph neural networks.

\subsection{Training Procedure for Lifelong Graph Learning}\label{sub:incrementaltraining}
Our incremental training algorithm for GNNs is shown in Algorithm~\ref{alg:proc}.
We assume to have a sequence of $T$ tasks $\gT_1,\ldots, \gT_T$ and a model $f$ with parameters $\theta$. 
Throughout the sequence of tasks, 
the graph changes in the sense that \nodes and edges are inserted and deleted. Crucially, the new \nodes can come with new classes that have not been part of the training data before.
To address these changes, we use the incremental training procedure from our prior work~\cite{galke2019graph} for adapting neural networks to new tasks.
As a preparation for task $\gT_{t}$, we retrain $f$ on the labels of $\gT_{t-1}$ to obtain $\theta^\embrace{t}$.
Whenever $l$ new classes appear in the training data, we add the corresponding number of parameters to the output layer of $f^\embrace{t}$. Therefore, we have $\lvert \theta^\embrace{t}_\text{output weights} \rvert  = \lvert \theta^\embrace{t-1}_\text{output weights}\rvert + l \cdot d_h$ and $\lvert \theta^\embrace{t}_\text{output bias}\rvert = \lvert \theta^\embrace{t-1}_\text{output bias}\rvert + l$, where $d_h$ is the output layer size.

These parameters that model the new classes are randomly initialized.
For the other parameters, we consider two options in our incremental training procedure: \emph{warm restarts} and \emph{cold restarts}.
With \emph{cold restarts}, we reinitialize $\theta^\embrace{t}$ and retrain from scratch.
On the contrary, when using \emph{warm restarts}, we initialize the parameters for training on task $\gT_t$ with the final parameters of the previous task $\theta^\embrace{t-1}$.
Furthermore, we incorporate a generic module (lines 12--14) for unseen class detection in the incremental training algorithm. This operates on the logits of the final output layer and determines whether the classification of a particular \node should be rejected because it belongs to a class that was not part of the training data.

\begin{algorithm}[h!]
\caption{Incremental training for lifelong graph learning under cold-start vs. warm-start condition (extended from~\cite{galke2021lifelong}). \label{alg:proc}}
\begin{algorithmic}[1]
  \Require{Sequence of tasks $\tilde{\gT}_0, \cdots, \tilde{\gT}_T$, model $f$ with parameters $\theta$, flag for cold or warm restarts
  \textbf{Output:} Predicted labels for new \nodes of each task along with decision whether it belongs to a previously known class 
  }

\State known\_classes $\leftarrow \emptyset$
  \State $\theta \leftarrow$ initialize\_parameters()
  \For{$t \leftarrow 1$ to $T$}  \Comment{Iterate through task indices}
  \State new\_classes $\leftarrow \operatorname{set}(\tilde{\vy}^\embrace{t-1}) \setminus \text{known\_classes}$
  \If{new\_classes $\neq \emptyset$}
    \State $\theta^\prime \leftarrow$ expand\_output\_layer($\theta$, $\lvert \text{new\_classes} \rvert$)
  \EndIf

  \State $\theta^\prime \leftarrow$ initialize\_parameters()
  \If{$t > 1$ \textnormal{\textbf{and}} do\_warm\_restart = \texttt{TRUE}} 
    \State $\theta^\prime \leftarrow$ copy\_existing\_parameters($\theta$) \Comment{Reuse prev. model}
  \EndIf

\State $\theta^\prime \leftarrow$ train($\theta^\prime$, $\tilde{\gG}_{t-1}$, $\tilde{\mX}^{(t-1)}$, $\tilde{\vy}^{(t-1)}$) 
  \Comment{Train model on prev. task}
  
  \State $\hat{\vy}_\text{logits}^{(t)} \leftarrow$ predict($\theta^\prime$, $\tilde{\gG_t}$, $\tilde{\mX}^{(t)}$) for $V_{t} \setminus V_{t-1}$ \Comment{Predict on new nodes}
  \State $\vm_\text{ood}^{(t)}= \operatorname{unseen\_class\_detection}(\hat{\vy}_\text{logits}^{(t)})$ \Comment{OOD-Detection}
  \State $\hat{\vy}_{\text{pred},i}^{(t)} = \begin{cases}
    \texttt{OOD} &\text{if}\, \vm_{\text{ood},i}^{(t)} = \texttt{TRUE} \\
    \argmax(\hat{\vy}_{\text{logits}^{(t)}, i}), & \text{otherwise}
 \end{cases}$
  \State known\_classes $\leftarrow \text{known\_classes} \cup \text{new\_classes}$ 
  \State $\theta \leftarrow \theta^\prime$
  \EndFor

\end{algorithmic}
\end{algorithm}

\subsection{Self-Detection of New Classes using our gDOC Method}\label{sub:methods:unseen}\label{gdoc}

\begin{figure}
    \centering
    \includegraphics[width=0.92\textwidth]{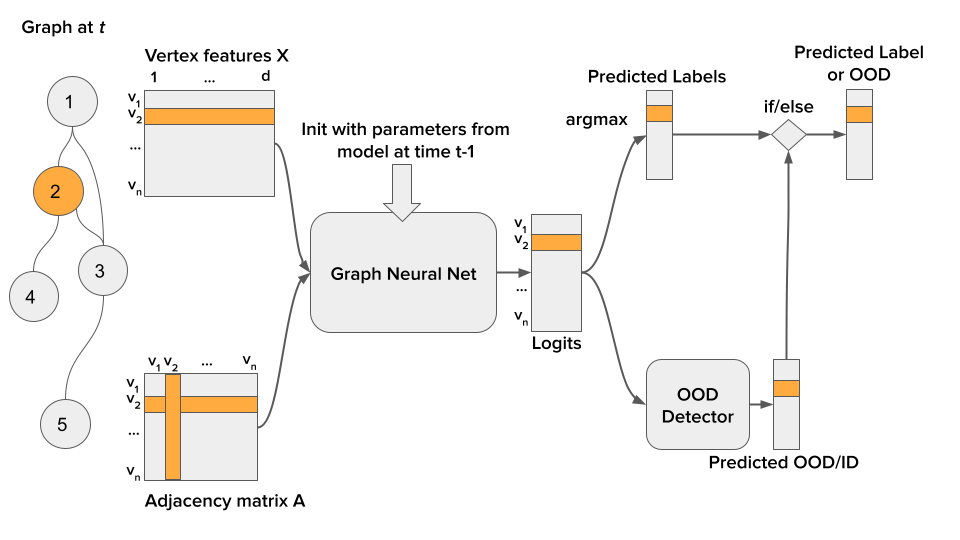}
    \caption{\revise{}{Procedure of node classification and OOD detection during the execution of a task in lifelong learning. 
    The output logits of the graph neural network are used in two ways. Once to determine the most likely in-distribution class and once to determine whether the example is in-distribution (belongs to a known class) or out-of-distribution. When an example is detected as in-distribution, we return the argmax of the logits. Otherwise, the example is marked as out-of-distribution.}}
    \label{fig:llg_architecture}
\end{figure}

A successful model for lifelong learning would not only classify new data into known classes but would also detect when an instance belongs to a previously unseen class.
We seek to develop a generic method that is not specific to any particular GNN architecture.
Thus, we take inspiration from the Deep Open Classification (DOC)~\cite{DOC} approach that has been proposed for text classification and transfer it to the graph domain.

\revise{}{The key challenges of transferring the DOC method from text to lifelong learning on graphs are how to deal with non-i.i.d. graph data and how to deal with an imbalanced class distribution. 
We tackle these challenges by combining DOC with a graph neural network and by weighting the binary cross-entropy loss function with the proportions of the class labels seen during training.}

\revise{}{\Figref{fig:llg_architecture} visualizes how we integrate an OOD detection module, such as DOC, into our lifelong node classification framework. 
A standard graph neural network emits logits for each \node, while the OOD detection module predicts whether a \node is in-distribution (ID) or OOD. If the OOD detector emits OOD, we reject to classify the respective \node with any of the known classes and assume that a new class has been observed. 
If the vertex is considered ID, the class label is assigned that corresponds to the maximum logit value.}

\revise{The}{To facilitate OOD detection, the} key idea \revise{}{of DOC} is to replace the final softmax activation with element-wise sigmoid activation.
Hence, the training objective becomes binary cross-entropy rather than categorical cross-entropy.
Then, thresholds on the logit distribution over all known classes are used to determine whether the new example belongs to an already known class or not.
Below, we briefly summarize the key risk reduction technique proposed in the original DOC, before we describe our extensions.

\paragraph{Thresholds and Risk Reduction in DOC}
To make a clear decision, a threshold is necessary to determine whether a vertex is considered out of distribution (OOD) at the test time.
When the output for all classes falls below the threshold, the classification of that vertex is rejected, \ie the vertex is considered OOD.
Such thresholds can be global or class-specific.
A natural choice for a global threshold $\tau$ is the inflection point of the sigmoid function, \ie setting $\tau = 0.5$.
However, estimating class-specific thresholds can further reduce the risk of incorrectly rejecting the classification of a known class. 
A strategy for estimating class-specific thresholds is to consult the standard deviation of logits in the training data~\cite{DOC}.
To determine a threshold $\tau_i$ for class $i$, the risk reduction technique proposed in DOC~\cite{DOC} collects all model outputs for instances of class $i$.
For all these outputs $\hat{y} \in [0,1]$, a mirror point $1+(1-\hat{y})$ is created, assuming a Gaussian distribution with mean 1.
On this distribution, the standard deviation $\operatorname{SD}_i$ is calculated to assign the class-specific threshold $\tau_i := \max \{\tau_\mathrm{min}, 1 - \alpha \cdot \operatorname{SD}_i\}$,
where $\alpha$ is a scaling factor for the standard deviation and $\tau_\mathrm{min}$ is the minimum threshold.
For $\alpha$, the original work suggests a value of $3$. The authors use a fixed $\tau_\mathrm{min} = 0.5$.
 
\paragraph{Extension to deal with class imbalance (gDOC)}
Here, we transfer the DOC method to the graph domain.
This comprises changing the base model from a 1D-CNN on text to a GNN operating on graphs, as well as changing the loss function for node classification from categorical cross-entropy to binary cross-entropy. In this way, we can employ the same strategy as the original work on DOC for detecting new classes. Throughout this work, we denote this adaptation from text to graph data as ``DOC''.

We propose an extension, which we denote as gDOC, to make DOC more suitable for lifelong learning on graphs, where we have to deal with a highly imbalanced class distribution.
We use a GNN model to emit the logits
and adjust the loss scaling of binary cross-entropy to account for class imbalance, which is inevitable in real-world graph data. 
This is particularly important for unseen class detection because here the magnitude of all outputs is relevant for the final decision, rather than only their maximum value. 
In detail, if class $i$ appears $n^+$ times in the training data, we multiply the loss of output $i$ by the factor $\frac{n - n^+}{n^+}$. This is a standard weighting procedure for binary cross-entropy that increases the loss according to the fraction of positive versus negative examples within the training data (\cf \cite{DBLP:journals/npl/AurelioACB19}).
We denote this variant as gDOC.
Furthermore, our experiments will carefully investigate different values for $\tau_\mathrm{min}$, while the original DOC~\cite{DOC} used a fixed minimum threshold of $\tau_\mathrm{min}=0.5$. Similarly, we also closely investigate the effect of the risk reduction factor $\alpha$.

\subsection{Summary}
We have extended the incremental training algorithm with a generic unseen class detection module.
As an unseen class detection module, we introduce gDOC as an extension of the DOC method from text to the challenges of lifelong learning with graph neural networks.
Note that both our adaptation of the original DOC to graphs (abbreviated simply as DOC) as well as our extended version (gDOC) can be employed in conjunction with arbitrary GNN base models.

\section{Measure of $k$-Neighborhood Time Differences}\label{sec:deltat}
Real-world graphs grow and change at different speeds~\cite{Aggarwal:2014:ENA:2620784.2601412}. 
Some graphs change quickly, such as social networks, while others evolve rather slowly, such as citation networks.
Furthermore, the graphs show different change behavior, \ie different patterns in how \nodes and edges are added and removed over time.
Therefore, depending on the specific graph data, a different history of the data must be used for training to take these factors into account.
To obtain absolute history sizes that are comparable across different temporal graphs, a measure is needed that provides a history size that is agnostic to the specific change dynamics of the graph (slow \vs fast).

Below, we first introduce such a measure, \revise{}{which we call $\Dt{k}$}.
In the experiments, we will use the $\Dt{k}$ measure to derive candidate history sizes as percentiles of the time differences in the data. Applying the measure can be regarded as a preprocessing step. 
Subsequently, we show that the history sizes that our measure produces are equivariant to the temporal granularity of the graph.

\subsection{Formal Definition of the $\Dt{k}$ Measure}

The $k$-neighborhood Time Difference Distribution measure $\Dt{k}$~\cite{galke2021lifelong} enumerates the distribution of time differences within the $k$-hop neighborhood of each \node.
This corresponds to the \emph{receptive field}~\cite{stochasticSampling} of a GNN with $k$-many graph convolutional layers.
Intuitively, we collect the time differences between all pairs of \nodes $v$ and $w$, which are reachable within at most $k$ edges.
We aggregate these time differences based on frequency, \ie we obtain the number of times a certain time difference has been observed between $v$ and $w$ in the dataset.
On this distribution of time differences (represented as a multiset), we compute the percentiles and use them as candidate history sizes.

\begin{definition}[$k$-Neighborhood Time Difference Distribution]
Given a graph $\gG = (V, E)$ and let $\mathcal{N}^k(u)$ be the $k$-hop neighborhood of vertex $u \in V$ with respect to $E$, \ie the set of \nodes that are reachable from $u$ by traversing at most $k$ edges. Let $\operatorname{time} : V \to \mathbb{N}$ be a function that returns the time information for each \node $v$ (timestamp metadata), \eg the year of publication when considering a citation graph.
We define $\Dt{k}(\gG)$ as \emph{multiset of time differences}, computed over all vertices $u \in V$ to their up to $k$-distant neighboring vertices $v \in \neighbors^k(u)$ that occurred before \node $u$. 

\begin{equation*}\label{eqn:deltat}
\Dt{k}(\gG) := \{ \tsmin(u) - \tsmin(v) \mid \forall u \in V \, \forall v \in \neighbors^k(u)
 \mathrm{~with~} \tsmin(v) \leq \tsmin(u) \}
\end{equation*}
\end{definition}

\begin{figure}
    \centering
    \includegraphics[width=0.65\linewidth]{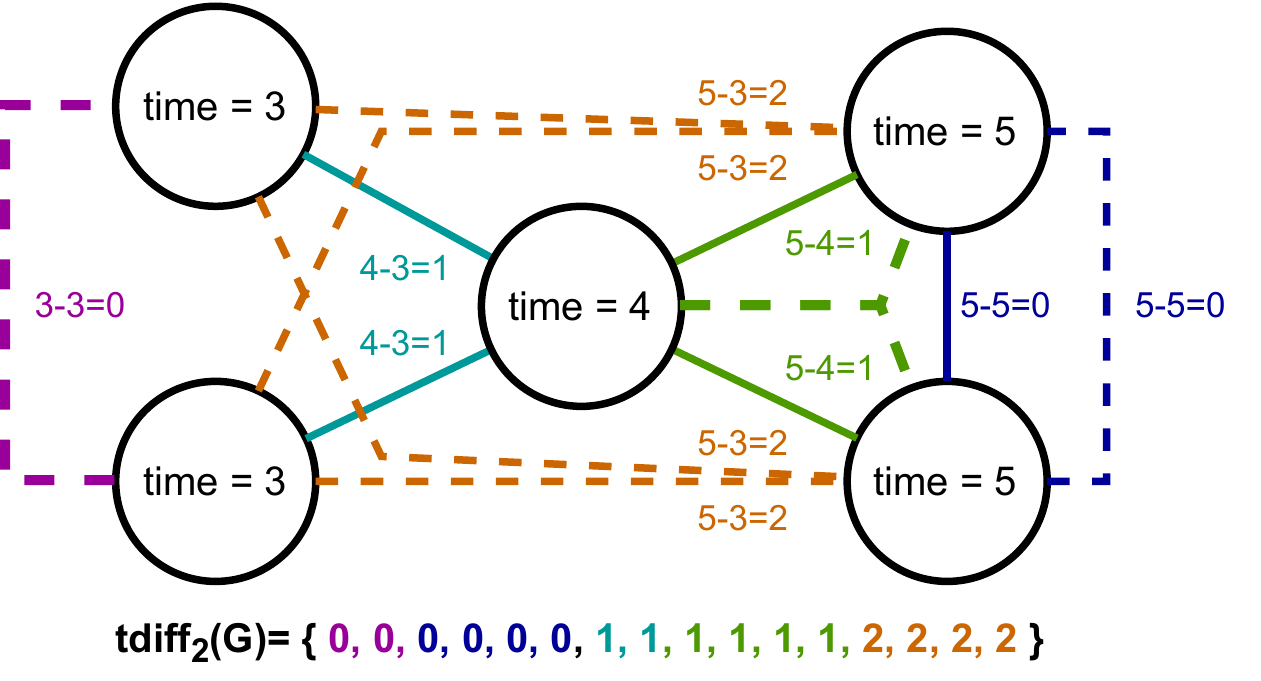}
    \caption{Example of time differences $\Dt{2}(G)$ for hops at distance of up to 2 from each \node. Solid lines are edges. Dashed lines indicate paths of length two. Annotations show the time difference between the endpoints of the path. The multiset $\Dt{2}(G)$ holds the resulting time differences. Note that zeros are counted in both directions as both fulfill the $\operatorname{time}(u) 
    \leq \operatorname{time}(v)$ condition.}
    \label{fig:tdiffk-example}
\end{figure}

The multiset $\Dt{k}$ maps each time difference to the respective number of occurrences and is interpreted as a distribution over time differences.
It is used to analyze the temporal distribution of the vertices in a dataset (using percentiles) and to make datasets comparable. 
Figure~\ref{fig:tdiffk-example} presents an exemplary computation of the $k$-neighborhood time differences $\Dt{2}$ on a graph with five \nodes and five edges. 
In this example, the 25th percentile of $\Dt{2}$ is $0$, the 50th is $1$ (also known as median), and the 
75th is also $1$.

The $\Dt{k}$ measure is used in our experiments to compare models trained with a \emph{limited history size} against models trained with the \emph{full history}.
Thus, we calculate the 25th, 50th, and 75th percentiles of the $\Dt{k}$ distribution, which we then compare against the full graph (100th percentile) to analyze the influence of explicit knowledge. 

\subsection{Equivariance to Temporal Granularity}
\newcommand{\ta}{t}
\newcommand{\tb}{t^\prime}
Any good measure to determine the discrete history sizes $c : (V, E, t) \mapsto \mathbb{N}$
in evolving graphs should be \emph{equivariant to granularity} to ensure comparability between different datasets and different granularities.
This means that if we change the perspective, for instance, from years to months, we should get history sizes that are about 12 times larger (on the same data).

More formally, consider two different time measurement functions $\ta,\tb : \gV \to \sN_{>0}$  whose values differ by a constant factor $a \in \sR^+$ such that $\ta(u) = \floor{\frac{\tb(u)}{a}} $ for all $  u\in \gV$.
For example, $a=12$ when comparing the granularities of months $\tb$ and years $\ta$.
In fact, two arbitrary discrete time measurement functions differ by a constant factor with one being coarser-grained (larger denominator) than the other or both being equal ($a=1$).
Then, the derived history sizes should not differ by more than the ratio between the granularity values, \ie for the measure $c$ to determine the history sizes it should hold that 
$c(V, E, t^\prime) \in \lbrack a \cdot c(V, E, t) - a; a \cdot c(V, E, t) + a \rbrack$
where $t^\prime \in [a \cdot t - a; a\cdot t + a]\,\text{~for all~} u \in V$. 
In the example above, a good measure $c$ should return a history size times 12 plus/minus 12 when we switch the perspective from the year level $t$ to the month level $t^\prime$ (ratio: $a=12$) on the same data. This property is also crucial for comparable history sizes across datasets with different temporal granularities.

Here, we briefly show that our $k$-neighborhood time difference measure $\Dt{k}$ is equivariant to temporal granularity:
We assume without loss of generality that $\tb$ is more fine-grained than $\ta$, \ie $a > 1$.
Because $\Dt{k}$ is a multiset of time differences from which we take percentiles to determine history sizes, it is sufficient to show that the time difference $t(u) - t(v)$ between two \nodes $u, v \in V$ is equivariant to the temporal granularity factor $a$, or more precisely: $\forall u,v \in V :\,  a \cdot \lvert \ta(u) - \ta(v) \rvert \in \left\rbrack \lvert \tb(u) - \tb(v) \rvert - a; \lvert \tb(u) - \tb(v) \rvert + a \right\lbrack$\revise{}{, where $]\cdot, \cdot[$ indicates an open interval}.

\paragraph{Prerequisite (PRE)}
With $\ta(u) = \left\lfloor {\frac{\tb(u)}{a}} \right\rfloor$
we have
$\frac{\tb(u)}{a} \leq \ta(u) < \frac{\tb(u) + a}{a}$
$\Rightarrow \tb(u) \leq a \cdot \ta(u) < \tb(u) + a$.
Note that the left-hand side is less than or equal due to rounding down, while the right-hand side adds a time step $a$, which makes it a true inequality.  

\begin{proof}
Using the prerequisite, we now show that
\begin{align*}
\text{(i)} \quad & a \cdot \lvert \ta(u) - \ta(v) \rvert > \lvert \tb(u) - \tb(v) \rvert - a \text{, and}\\
\text{(ii)} \quad & a \cdot \lvert \ta(u) - \ta(v) \rvert < \lvert \tb(u) - \tb(v) \rvert + a  
\end{align*}
via case differentiation.

\paragraph{Case (i-a): $\ta(u) = \ta(v)$} The left-hand side of inequality (i) becomes zero and it remains to show that $\lvert \tb(u) - \tb(v)\rvert  < a$. 
We apply (PRE) to find the highest possible value for the term $\lvert \tb(u) - \tb(v)\rvert$ with respect to $\ta$ such that the term is still smaller than $a$.
The highest possible value for $\tb(u)$, expressed in terms of $\ta$, is $a\cdot \ta(u) + \epsilon$ with $0 < \epsilon < a$. This is because $a \cdot \ta(u)$ is the upper bound of $\tb(u)$ in the inequality of the prerequisite (PRE) and we insert a small but positive $\epsilon$ to account for ``truly lesser''.
The smallest possible value for $\tb(v)$ is $a\cdot \ta(u)$.
Again, we take this value $a \cdot \ta(u)$ from the prerequisite, where it is the lower bound for $\ta(u)$.
Then, we have $\lvert a \cdot \ta(u) + a - \epsilon - a \cdot \ta(v)\rvert < a$. Now, as $\ta(u) = \ta(v)$, we obtain $\lvert a-\epsilon \rvert < a$.

\paragraph{Case (i-b): $\ta(u) \neq \ta(v)$} We transform the left-hand side of (i) to $\lvert  a \cdot \ta(u) - a \cdot \ta(v)\rvert $, while recalling that $\ta \in \sN_{>0}$. 
We use (PRE) to obtain $\lvert \tb(u) - \tb(v)\rvert $ as the smallest possible value on the left-hand side. 
Now, the left and right sides are the same except for $-a$ on the right. 
As $a > 1$ (is positive), inequality (i) is valid. 

\paragraph{Case (ii-a): $\ta(u) = \ta(v)$} The left-hand side of (ii) becomes zero and it remains to show that $0 < \lvert \tb(u) - \tb(v)\rvert  + a$, which is true because $a>1$.

\paragraph{Case (ii-b): $\ta(u) \neq \ta(v)$} Again, we transform the left-hand side of (ii) to $\lvert  a \cdot \ta(u) - a \cdot \ta(v)\rvert $. This time, we are interested in the highest possible value with respect to (PRE), which is $\lvert \tb(u) + a - \epsilon - \tb(v) \rvert $ with $0 < \epsilon < a$.
This is because the highest possible difference is between the upper bound $\tb + a - \epsilon$ and the lower bound $\tb$.
With the triangle inequality, we obtain $\lvert \tb(u) + a - \epsilon - \tb(v)\rvert  \leq \lvert \tb(u) - \tb(v) \rvert + \lvert a - \epsilon \rvert < \lvert \tb(u) - \tb(v) \rvert + a$,
which holds because $\lvert a - \epsilon \rvert < a$.
This concludes the proof.
\end{proof}

\subsection{Summary}

The $k$-neighborhood Time Difference Distribution $\Dt{k}$ measures the granularity and temporal connectivity patterns of the given graph dataset with \node{}-level time information. 
In general, we can hardly assume that any absolute history size on dataset~A would be comparable to the same history size on dataset~B. But if we derive the history size from $\Dt{k}$, \eg the median of $\Dt{2}$, we have a strategy to find comparable history sizes across datasets, even if they come from different domains, 
\eg social graphs with postings at the minute level versus citation graphs with data on, at least, daily level. 
This is because our $\Dt{k}$ measure is equivariant to granularity and is based solely on time differences between the connected \nodes.

\section{Datasets and Analyses}\label{sec:datasets}
Adapting models to new data is an important problem whenever machine learning models are deployed in production.
However, many graph benchmark datasets are stripped of any temporal data, which is needed to divide the data into realistic partitions for lifelong learning, \ie a sequence of tasks.
To build a lifelong vertex classification dataset with new classes, the following criteria need to be fulfilled:

\begin{itemize}
    \item attributed \nodes,
    \item \node labels,
    \item time information on the \nodes,
    \item \emph{evolving} set of \nodes (and thus also edges) over time,
    \item \emph{evolving} set of classes over time, especially the occurence of new classes.
\end{itemize}

We scan the literature (\eg \cite{dwivedi2020benchmarking,DBLP:conf/aaai/ParejaDCMSKKSL20,Xu2020Inductive}) and common dataset collections (OpenGraphBenchmark~\cite{hu2020open}, KONECT\footnote{\url{http://konect.cc/}}, and PyTorch Geometric Temporal\footnote{\url{https://pytorch-geometric-temporal.readthedocs.io/}}) for datasets that met the criteria above.

Surprisingly, preprocessed graph datasets that meet these criteria are rare, even though the raw origin of these datasets (social media data, publication metadata) would meet the requirements. In those datasets, in which time information is available, either the graph is static, \ie it is not an evolving graph, or the set of classes is static, \ie there are no newly appearing classes over time. Concurrent work on lifelong learning composes the sequence of tasks by synthesizing an ordering of the \nodes in static graph datasets~\cite{wangLifelongGraphLearning2021}, \ie data-incremental or class-incremental experimental setups.

In this work, we seek to understand how our methods deal with real-world datasets, \ie we simulate the evolution of the graph along the time axis and add new \nodes and edges according to the time stamps of the \nodes.
For our first experiment, we used two different splits on standard benchmark datasets with static graphs, which are described next. 
We can use these datasets to simulate two steps of a temporal graph, where the training data is step one and the \emph{unlabeled} test data is step two.
Thereafter, we describe our own temporal datasets that we contribute to the community and use for the other three experiments on lifelong learning.

\subsection{Static Graph Datasets}\label{sub:standard-datasets}

Cora, Citeseer, and PubMed are standard datasets for the vertex classification task~\cite{DBLP:journals/aim/SenNBGGE08},
which we use for our first experiments on transductive versus inductive learning.
The \nodes of the graph are research articles represented by textual features and
annotated with a class label. Edges are defined by citation relationships but are considered bidirectional. These datasets are often used in transductive
learning environments~\cite{DBLP:conf/icml/YangCS16,DBLP:journals/corr/KipfW16,velickovic2018graph}.
In our experimental setup, we use these datasets to compare inductive \vs transductive learning.
\begin{table}[t]
  \caption{Statistics for train-test splits: few-many (A) and many-few (B) settings on the citation networks
    datasets: Cora, Citeseer, and Pubmed. The unseen \nodes and edges are available
    only after the training epochs. The test samples for measuring accuracy are
  a subset of the unseen \nodes. The label rate is the percentage of labeled \nodes for training.}
  \label{tab:datasets:motivational}
  \begin{center}
    \begin{tabular}{lrrrrrr}
    \toprule
      \textbf{Dataset} & \multicolumn{2}{r}{\bf Cora}  &\multicolumn{2}{r}{\bf Citeseer} & \multicolumn{2}{r}{\bf Pubmed} \\
      \midrule
      Classes & \multicolumn{2}{r}{7} & \multicolumn{2}{r}{6} & \multicolumn{2}{r}{3}\\
      Features & \multicolumn{2}{r}{1,433} &\multicolumn{2}{r}{3,703} & \multicolumn{2}{r}{500}\\
      \Nodes & \multicolumn{2}{r}{2,708} & \multicolumn{2}{r}{3,327} & \multicolumn{2}{r}{19,717}\\
      Edges & \multicolumn{2}{r}{5,278} & \multicolumn{2}{r}{4,552} & \multicolumn{2}{r}{44,324}\\
      Avg. Degree & \multicolumn{2}{r}{3.90} & \multicolumn{2}{r}{2.77} &
      \multicolumn{2}{r}{4.50}\\
      \midrule
      \textbf{Setting} & \textbf{A} & \textbf{B} & \textbf{A} & \textbf{B} & \textbf{A} & \textbf{B} \\
      \midrule 
      Train \Nodes & 440        & 2,268      & 620        & 2,707      & 560        & 19,157     \\
      Train Edges    & 342        & 3,582      & 139        & 2,939      & 34         & 41,858        \\
      Unseen \Nodes   & 2,268      & 440        & 2,707      & 620        & 19,157     & 560     \\
      Unseen Edges   & 4,936      & 1,696      & 4,413      & 1,613      & 44,290     & 2,466     \\
      Test Samples   & 1,000      & 440        & 1,000      & 620        & 1,000      & 560        \\
      Label Rate     & 16.2\%      & 83.8\%      & 18.6\%      & 81.4\%      & 2.8\% & 97.2\%      \\
      \bottomrule
    \end{tabular}
  \end{center}
\end{table}

We prepare the static graph datasets in two ways: either a lot of training data and a few test data, or vice versa.   Specifically, we used two different train-test splits for each dataset, which we call setting $A$ and setting $B$.
The setting $A$ is derived from the original train-test split for transductive
tasks~\cite{DBLP:journals/corr/KipfW16}. It consists of a few labeled \nodes that
induce our training set and many unlabeled \nodes.
Setting B instead comprises many training \nodes and few test \nodes. We set it up
by inverting the train-test mask of Setting A and assigning the edges accordingly.
Setting B is motivated by applications, in which a large graph is already known and incremental changes occur over time, such as for citation recommendations, link prediction in social networks, and others~\cite{Aggarwal:2014:ENA:2620784.2601412,Galke:2018:MAA:3209219.3209236}. 
We refer to Table~\ref{tab:datasets:motivational} for the details of the datasets and the two
settings. We used these three data sets with two different train-test splits in our first experiment described in \Secref{exp:transductive-inductive}.

\subsection{Evolving Graph Datasets}\label{sub:new-datasets}
We published three graph datasets for lifelong learning~\cite{galke2021lifelong}: one co-authorship graph dataset
(\wos) and two DBLP-based citation graph datasets (\dblpeasy and \dblphard).
For \wos, the classes are journal categories.
For DBLP, we use conferences and journals of published papers as classes.
Since we select those venues with the most publications, this serves as a proxy for a broad categorization.
When new conferences and journals emerge, as they do in computer science, new classes appear in the data.
 
The datasets were generated by imposing a minimum threshold of publications per class per year: 100 for \dblpeasy, 45 for \dblphard, and 20 for \wos.
For the co-authorship graph \wos{}, we additionally require a minimum of two publications per author per year.
In all datasets, \node features are normalized TF-IDF representations of the publication title.

\begin{table}[!th]
    \centering
    \caption{Global dataset characteristics: total number of \nodes $|V|$, edges $\lvert E \rvert$, features $D$, and classes $\lvert \sY \rvert$ along with number of newly appearing classes (in braces) within the $T$ evaluation tasks}\label{tab:datasets}
    \begin{tabular}{lrrrrrr}
    \toprule
    Dataset & $\lvert V \rvert$ & $\lvert E \rvert$ & $D$ & $\lvert \sY \rvert$ & $T$\\
    \midrule
    \dblpeasy & 45,407  & 112,131 & 2,278 & 12 (4 new)    & 12\\
    \dblphard & 198,675 & 643,734 & 4,043 & 73 (23 new)  & 12\\
    \wos      & 68,068  & 2,1M    & 4,829 & 7 (0 new)  & 18\\
    \bottomrule
    \end{tabular}
\end{table}
\begin{figure*}[t]
    \centering
    \includegraphics[width=0.32\textwidth]{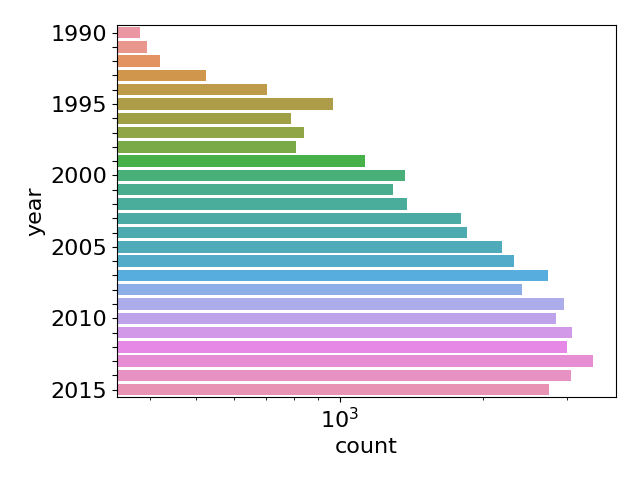}
    \includegraphics[width=0.32\textwidth]{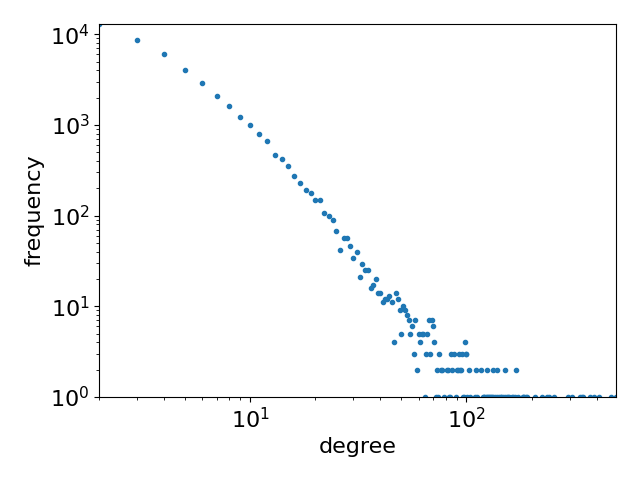}
    \includegraphics[width=0.32\textwidth]{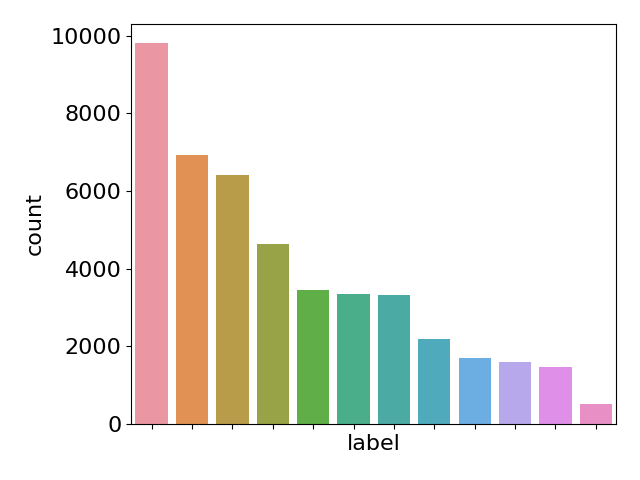}
    \includegraphics[width=0.32\textwidth]{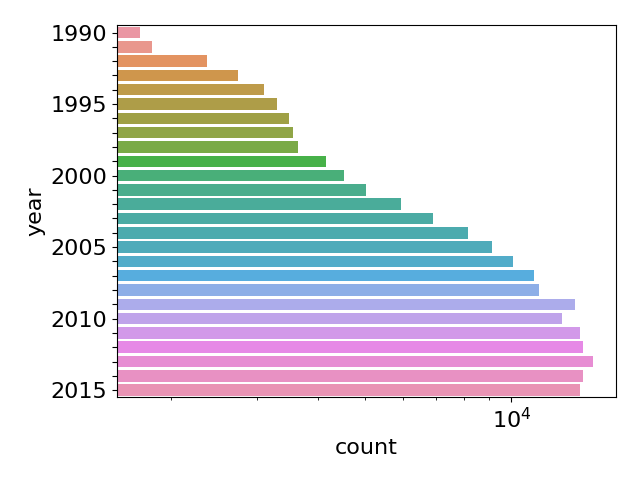}
    \includegraphics[width=0.32\textwidth]{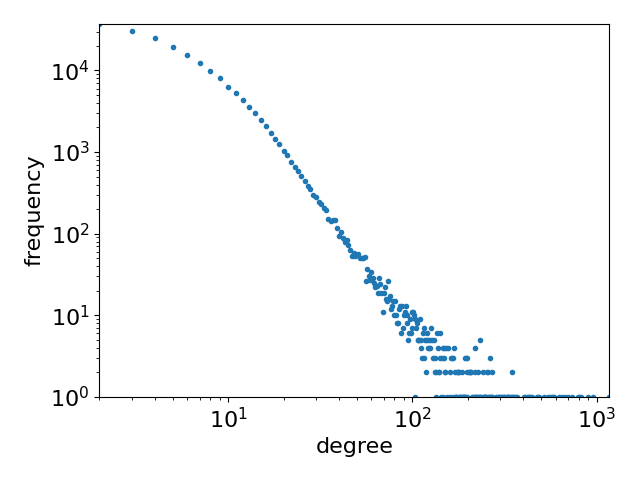}
    \includegraphics[width=0.32\textwidth]{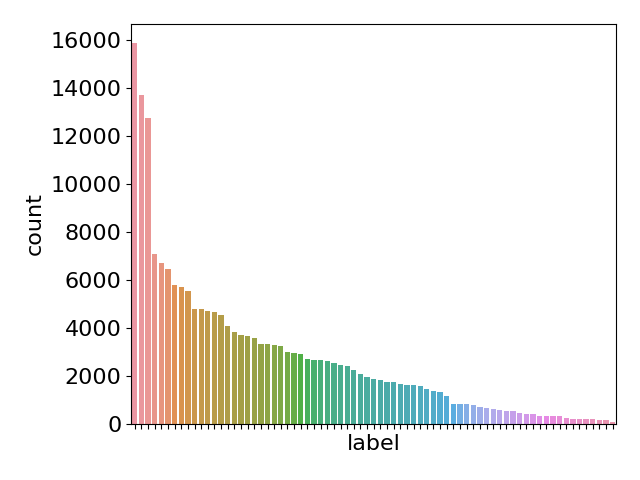}
    \includegraphics[width=0.32\textwidth]{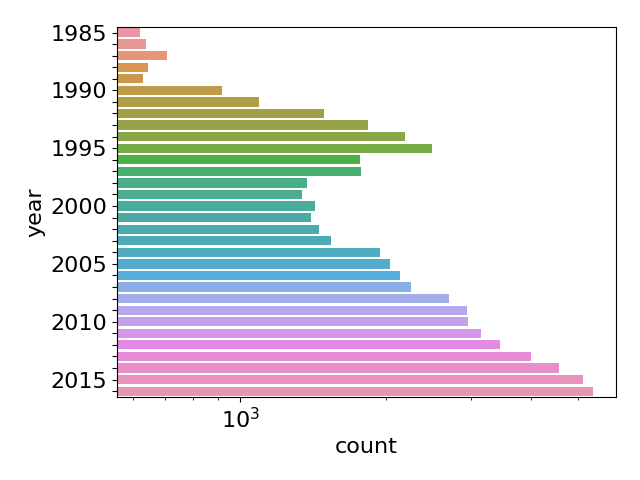}
    \includegraphics[width=0.32\textwidth]{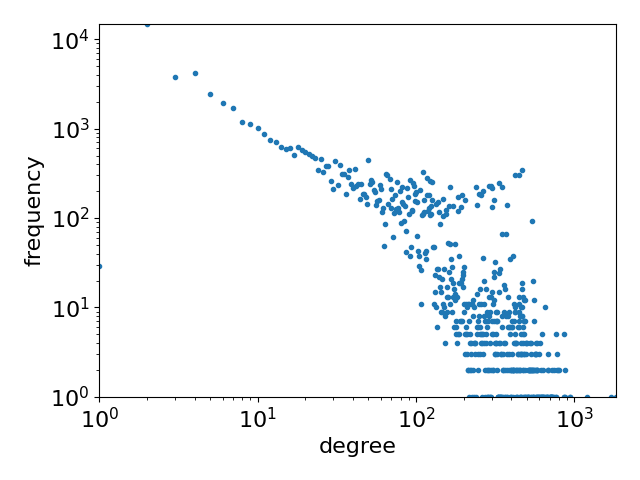}
    \includegraphics[width=0.32\textwidth]{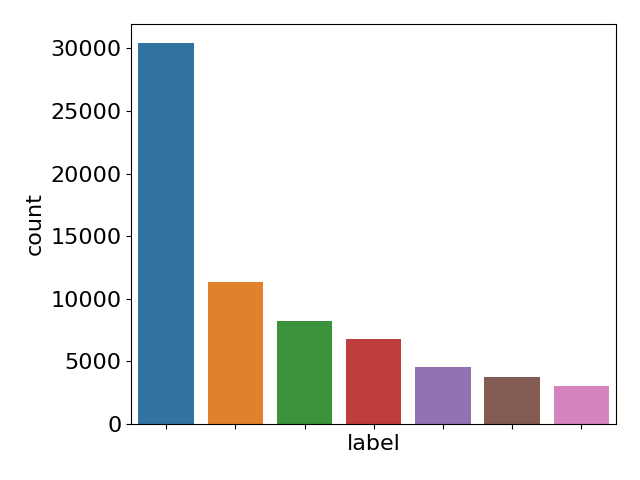}
    \caption{Distribution of vertices per year on log scale (left column), degree distributions (middle column), label distributions (right column), for our new datasets: \dblpeasy (top row), \dblphard (middle row), \wos (bottom row)}\label{fig:datasets}
\end{figure*}

\subsubsection{Basic Characteristics}
Table~\ref{tab:datasets} summarizes the basic characteristics of the datasets.
\dblpeasy and \dblphard are organized into 12 annual snapshots, while \wos has 18 annual snapshots.
\dblpeasy has 45k \nodes, 112k edges, and a feature dimension of 2,278. The \nodes are assigned to one of 12 classes, of which four only appear during the sequence of snapshots, \ie they are not present in the first snapshots.
\dblphard has 199k \nodes, 644k edges,  and a feature dimension of 4,043 (because the word vocabulary is set up based on occurrences within documents). Twenty-three of the 73 classes appear only during subsequent snapshots. \wos comes with 68k \nodes, 2.1M edges, feature dimension 4,829, 7 classes, and 18 snapshots. The number of edges is much higher than in the DBLP variants because \wos is a coauthorship graph, which is denser than the citation graphs.
Note that \dblpeasy is a subset of \dblphard as both were generated by applying a minimum threshold on the number of publications per class. 

We report the label distribution of the datasets, the degree distribution, and the distribution over time in \Figref{fig:datasets}.
The annual number of publications grows over time.
Only in \wos, there is a higher amount of \nodes between 1991-1997 than between 1998 and 2003.
The global degree distributions of \dblpeasy and \dblphard seem to follow a power-law distribution~\cite{newman2005power} as the degree distribution is almost a straight line except for the blurry tail.
For \wos, the degree distribution is more blurry, while a trend line can still be identified.
Furthermore, we observe that the number of examples per class is imbalanced in all three datasets.
Although the three datasets have different numbers of classes, the shape of the label distributions is similar.

\subsubsection{Changes in the Class Set and Distribution Shift}
Regarding changes in the set of classes, \dblpeasy has 12
venues in total, including one biannual conference and four venues, which appear in 2005, 2006, 2007, and 2012. \dblphard has 73 venues,
including one discontinued, nine biannual, six irregular venues, and 23 new venues.
To quantify the changes in the class set, we calculate the magnitude of the class drift as the total variation distance~\cite{webb2016,webb2018}:
\begin{equation*}
    \sigma_{t-1,t} = \frac12 \sum_{y \in \sY_{t-1} \cup \sY_t} \lvert P_{t-1}(y) - P_{t}(y) \rvert 
\end{equation*}
where $P_t(y)$ is the observed class probability at time $t$. 
\begin{figure}[ht]
    \centering
    \includegraphics[width=0.95\linewidth]{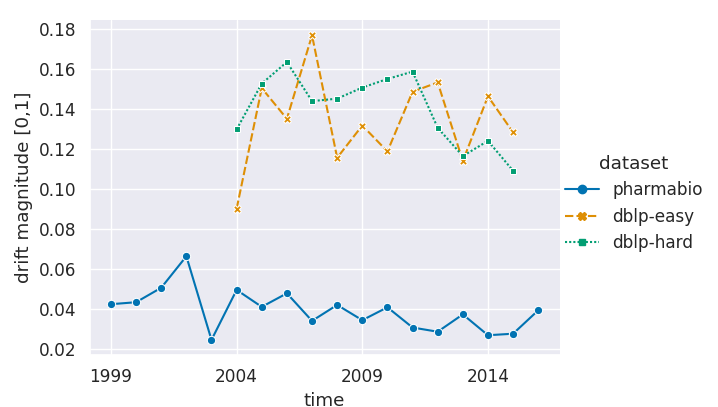}
  \caption{Magnitude of the class drift per dataset. The drift within the \wos dataset (no new classes) is lower than the drift of both DBLP variants. Independent and identically distributed data would have drift magnitude zero.}
    \label{fig:drift_mag}
\end{figure}
We visualize the drift magnitudes per dataset in \Figref{fig:drift_mag}. An IID dataset would have a drift magnitude of zero by definition. As expected, the drift magnitude is high (between 0.12 and 0.16)
 for the two datasets with new classes: \dblpeasy and \dblphard.
On \wos, which does not have new classes, the drift magnitude is consistently lower than 0.07.

\subsubsection{Analyzing Time Differences using $\Dt{k}$}
\begin{figure*}[ht]
    \centering
    \includegraphics[width=0.95\linewidth]{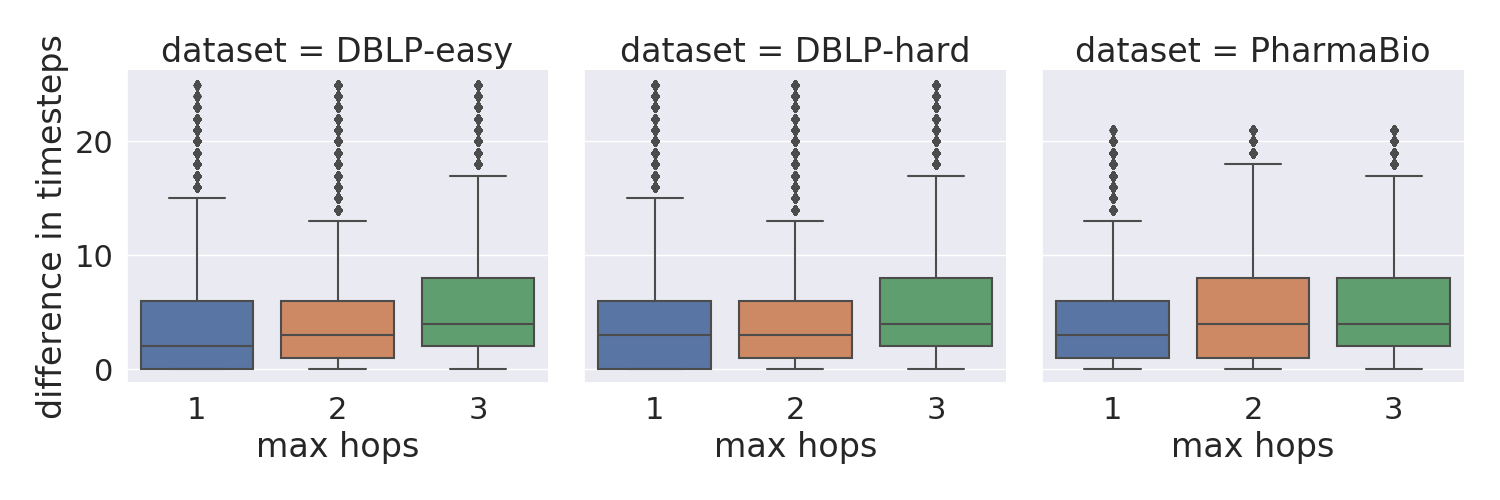}
    \caption{Distributions of time differences $\Dt{k}$ (y-axis) for \dblpeasy (left), \dblphard (center) and \wos (right) within the $k$-hop neighborhood for $k = \{ 1,2,3\}$ (x-axis).}\label{fig:deltat}
\end{figure*}

We analyze each dataset using our $k$-neighborhood time differences $\Dt{k}$ introduced in \Secref{sec:deltat}.
In \Figref{fig:deltat}, we show the distributions for three different values of $k=1,2,3$.
As expected, the time differences increase if we allow a longer maximum path length $k$.
For our experiments, we will use GNN models with 2 layers, \ie, which take into account the two-hop neighborhood of each \node.
Thus, we use $\Dt{2}$ to derive candidate history sizes, which we will compare to each other in the experiments.
Following the distributions for $k=2$ depicted in Figure~\ref{fig:deltat}, we select $1$, $3$, $6$, and $25$ as history sizes for DBLP-\{easy,hard\} and $1$, $4$, $8$, and $21$ as history sizes for \wos according to the $25$th, $50$th, $75$th, and $100$th percentiles of $\Dt{2}$.

\subsubsection{Dataset Preprocessing}
For each dynamic dataset, we construct the sequence of tasks $\tilde{\gT}_1, \ldots,\tilde{\gT}_T$ based on the publication year along with a history size $c$.
For each task $\tilde{\gT}_t$, we construct a graph with publications of time $\lbrack t - c, t \rbrack$, where publications of time $t$ are the test \nodes, and $t < c$ training \nodes (transductive).
We also use it for inductive training, where we train exclusively on $\tilde{\gT}_{t-1}$, but still evaluate the test \nodes of $\tilde{\gT}_{t}$

We set the first evaluation task $\tilde{\gT}_1$ to the time,
at which 25\% of the total number of publications are available.
Therefore, by mapping the datasets to our problem statement (see Figure~\ref{fig:teaser}), our first evaluation task $t=1$ corresponds to the year 1999 in \wos (1985--2016) and 2004 in DBLP-\{easy,hard\} (1990-2015). We continue to iterate over the next years for subsequent tasks, \ie from 2000 to 2016 for PharmaBio and from 2005 to 2015 for DBLP.

\subsection{Summary}
We have three static graph datasets (Cora, Citeseer, and Pubmed) and three dynamic graph datasets (PharmaBio, DBLP-easy, and DBLP-hard). All datasets have scientific publications as \nodes. 
All datasets use citations to set up the edges of the graph, except PharmaBio, where the edges are determined by coauthorships.
All graph datasets have a highly imbalanced label distribution (see Figure~\ref{fig:datasets}).
Two of the dynamic graph datasets come with new classes: \dblpeasy{} and \dblphard{},
which is reflected in a high class drift over time (see Figure~\ref{fig:drift_mag}).

We will use the static graph datasets in the first experiments described in Section~\ref{exp:transductive-inductive}.
We will use the dynamic graph datasets in the experiments described in Sections \ref{exp:lifelong-learning}, \ref{exp:limited-labeled-data}, and \ref{exp:open-learning}.

\section{Experiment 1: Transductive versus Inductive Learning}\label{exp:transductive-inductive}
In the first experiment, our objective is to learn whether accuracy increases when we add unlabeled data to the graph after having trained a model only on the portion of the graph that has labeled vertices.
This is important for later experiments because it affects how we move from task $t$ to task $t+1$. 
We answer whether we need to retrain a model with unlabeled data from the graph at $t+1$, or is it sufficient to wait until the new labeled data become part of the training set. 
This research question can be very well investigated with the static graph datasets that we introduced in \Secref{sub:standard-datasets}.
We use the training set of the static graphs as step one and the unlabeled part of the test set as step two.
In order to obtain generalizable results, we consider two different train-test splits for each dataset, which we call setting $A$ (few training, many test examples) and setting $B$ (many training, few test examples), as described in more detail in \Secref{sub:standard-datasets}.

In the context of lifelong learning, settings A and B correspond to different stages of the incremental training procedure. At the very beginning, we start with a few labeled data. After a few tasks, the number of labeled vertices increases, and, then, any new data added to the training set will make only a smaller fraction of the already known labeled data.

In the following, we describe the procedure, hyperparameter, and metrics of our experiments to analyze transductive vs. inductive learning on standard benchmark datasets with two complementary train-test splits. 
The aim is to analyze the effect of adding unlabeled data after (pre-)training and comparing inductively pre-trained models to models that have been trained transductively \emph{including} the unlabeled test data. 
We will show that the addition of unlabeled data does not further improve the performance of the inductively pre-trained models.

\subsection{Procedure}
We construct a dedicated experimental setup to assess the inference
capabilities of graph neural networks. We include edges in the training set if
and only if its source and destination \node are both in the training set.
The training process is then divided into two steps. First, we pre-train the model on
the labeled training set. Then we insert the previously unseen
\nodes and edges into the graph and continue training for a limited number of
inference epochs. The unseen \nodes do not introduce any new labels. Instead, the
unseen \nodes provide features and may be connected to known labeled \nodes. We
evaluate the accuracy on the test \nodes, which form a subset of the unseen \nodes,
before the first and after each inference epoch. 
We consider the graph neural networks GCN, GAT, GraphSAGE, as discussed in \Secref{sec:rw}, along with a baseline model MLP.
For each model, we compare 200 pre-training epochs versus no pre-training. In the latter case,
training begins during inference, which is equivalent to retraining from
scratch whenever new \nodes and edges are inserted. This allows us to assess
whether pre-training is helpful for applying graph neural networks on dynamic
graphs.

\subsection{Hyperparameters}
All employed graph neural networks use two graph convolution layers that aggregate neighbor
representations.  The output dimension of the second layer corresponds to the number of classes.
Thus, the features within the two-hop neighborhood of each labeled \node are
taken into account for its prediction.
We adopt the same hyperparameter values as proposed in the respective original
work. For GCN, we use 16 or 64 hidden units (denoted GCN-64) per layer, ReLU
activation, 0.5 dropout rate, along with an (initial) learning rate
of 0.005 and weight decay $5 \cdot
10^{-4}$~\cite{DBLP:journals/corr/KipfW16}.
For GAT, we use 8 hidden units per layer and 8
attention heads in the first layer. The second layer has 1 attention head (8 on
Pubmed). We set the learning rate to 0.005 (0.01 on Pubmed) with weight decay
0.0005 (0.001 on Pubmed)~\cite{velickovic2018graph}. 
For GraphSAGE, we use 64 hidden units per layer with mean aggregation, ReLU
activation, and a dropout rate of 0.5. 
We set the learning rate to 0.01 with weight decay $5 \cdot
10^{-4}$~\cite{DBLP:conf/nips/HamiltonYL17}.
Our MLP baseline has one hidden layer with 64 hidden units, ReLU
activation, a dropout rate of 0.5, a learning rate of 0.005 and a weight decay of $5 \cdot
10^{-4}$.
In all cases, we use  
Glorot initialization~\cite{DBLP:journals/jmlr/GlorotB10} and Adam~\cite{Adam} to optimize cross-entropy.
We initialize the optimizer at the beginning of the inference epochs.

\subsection{Measures}
\paragraph{Accuracy}
We train each model for 35 epochs and repeat the training 100 times with different seeds.
The plot shows the mean accuracy plus the standard deviation of the models at each of these training epochs.

\paragraph{Jenson-Shannon Divergence}
We further compute the Jenson-Shannon divergence~\cite{DBLP:journals/tit/Lin91} on the accuracy distributions to quantify the similarity of the distributions in the two different pre-training configurations (with or without) and in the two different settings (A and B). Since the two distributions are of the same kind, we use a symmetric measure to compare them.

The Jenson-Shannon divergence ($D_\mathrm{JS}$) is such a symmetric measure. It compares two distributions $P$ and $Q$ by calculating the (asymmetric) Kullback-Leibler divergence ($D_\mathrm{KL}$) in both directions:

\begin{equation*}
    D_\mathrm{JS}(P||Q) = \frac12 D_\mathrm{KL}(P || Q) + \frac12 D_\mathrm{KL}(Q || P)
\end{equation*}
As $D_\mathrm{JS}$ is a divergence measure, lower values indicate more similar distributions.

\subsection{Results}
\begin{figure*}[ht] 
  \centering
  \includegraphics[width=0.98\linewidth]{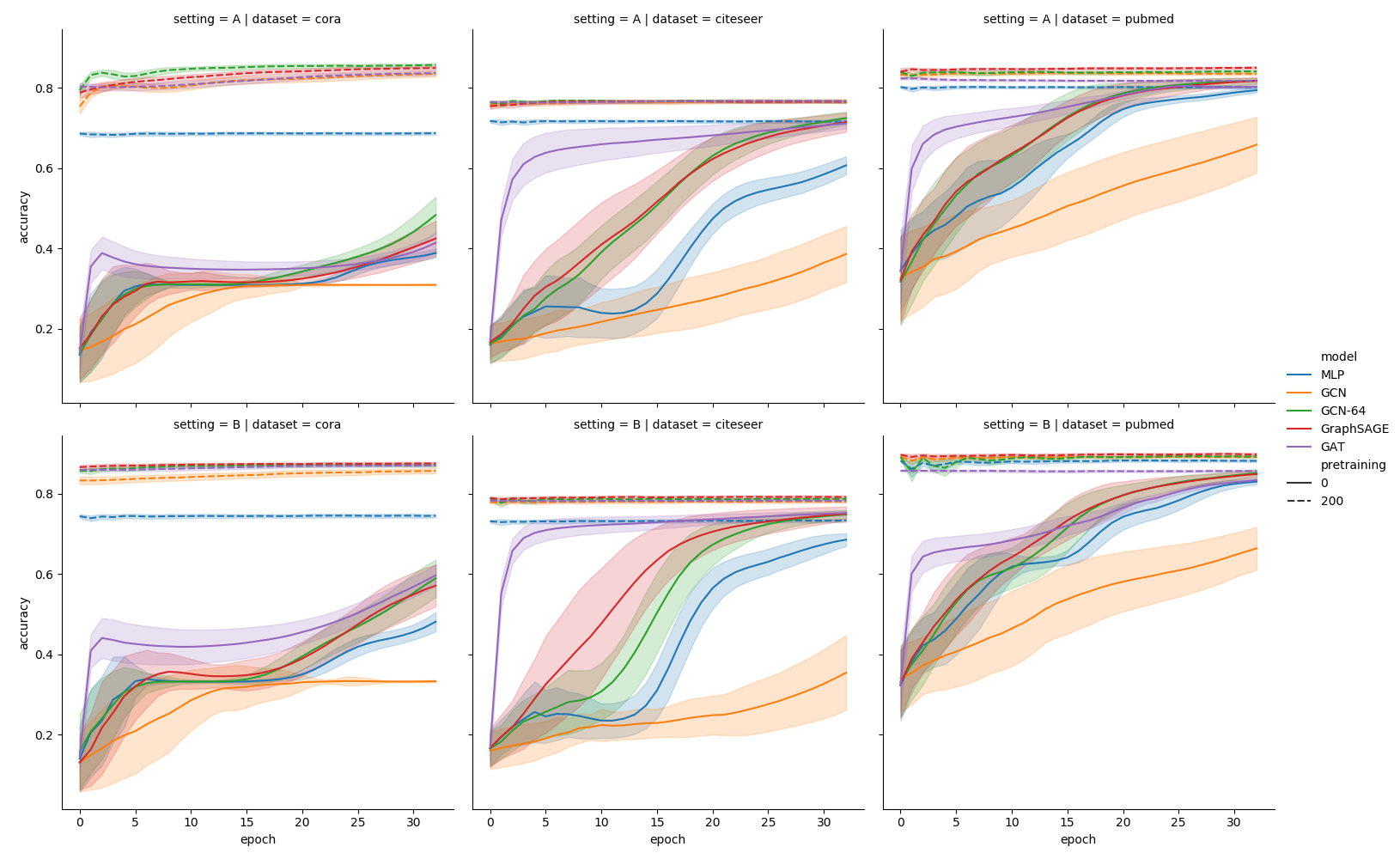}
  \caption{Test accuracy after each inference epoch for the many-few settings A \figtop{}
    and few-many setting B \figbottom{} on the datasets Cora, Citeseer, and Pubmed. Each line resembles the mean
    of 100~runs and its region shows the standard deviation.
    The dashed lines show the results with 200~pre-training. The solid lines are the results without pre-training.
  }\label{fig:results}
\end{figure*}

\Figref{fig:results} shows the results of the GNN models and the MLP on the
three datasets: Cora, Citeseer, and Pubmed.
The scores of the many-few setting B are higher than those of the few-many setting A by a
constant margin.
Pre-trained models score consistently
higher than non-pre-trained models while having less variance.
The accuracy of the pre-trained models plateaus after a few inference epochs (up to 10 on
Cora-A, \ie the Cora dataset investigated in setting A, and Pubmed-B, \ie setting B applied on the Pubmed dataset).
Without any pre-training, GAT shows the fastest learning process. 
The absolute scores of pre-trained graph neural networks are higher than the
ones of MLP. From a broad perspective, the scores of pre-trained graph neural
networks are all on the same level. While GCN falls behind the others on Cora-B,
GAT falls behind the others on Pubmed in both settings.

We compare the results of setting A and B by measuring the
Jensen-Shannon divergence between the accuracy distributions.
The Jenson-Shannon divergence between the two settings is lower with
pre-training (between 0.0057 for GAT and 0.0115 for MLP) than it is without
pre-training (between 0.0666 for GraphSAGE and 0.1013 for GCN).
This shows that the accuracy distributions are similar in both train-test
splits.

\subsection{Summary}
Our results show that inductive graph neural networks perform well even though we insert new
\emph{unlabeled} \nodes and edges after training.
For all three datasets, the accuracy plateaus after
very few inference epochs. 
This observation holds for both train-test split settings A and B, \ie many-few and few-many data for training and testing, respectively.
In different terms, we have not observed any gain from up-training an inductive model on extra \emph{unlabeled} data.
This motivates us to use the warm restart strategy, \ie reusing previous parameters, in the following experiments on lifelong learning. 

\section{Experiment 2: Lifelong Learning on Graphs}\label{exp:lifelong-learning}
From the previous experiment, we know that inductively trained models are stable when adding unlabeled data after training. 
Now, we focus on the case in which we continually add more labeled data to the graph, even including new classes in addition to new \nodes and edges.
The aim is to determine whether parameter reuse is helpful. 
We consider this question in the context of whether and how many old \nodes (and their edges) can be discarded when dealing with evolving graphs.

The challenge in this experiment is that the GNN models have to sequentially adapt to new tasks with new labeled data including unseen classes. 
We apply the GNNs GraphSAGE, GAT, SGC, GraphSAINT, JKNet, and the baseline MLP on our evolving graph datasets, which we described in \Secref{sub:new-datasets}.
As we know from our analyses of \Secref{sec:datasets}, the dynamic datasets are naturally heavily imbalanced. The datasets also feature new classes that appear over time. The first appearance of a new class is always at test time, and, only afterward, the \nodes with new classes are only added to the training data.
In summary, we find interactions between implicit and explicit knowledge:
Reusing past parameters (warm restart) enables using smaller history sizes with
only a small decrease in performance.

\subsection{Procedure}
 The evolving graph is divided into tasks according to the time slices in years (see Section~\ref{sub:new-datasets}).
 We apply the incremental training algorithm of Section~\ref{sub:incrementaltraining} to each of the considered models, GraphSAGE, GAT, SGC, GraphSAINT, JKNet, along with a graph-agnostic MLP baseline.
 The rationale for the selection of these particular base GNN models is provided in \Secref{selection-of-base-models}.
 
 For each model, we distinguish between warm restart and cold restart configurations, which determines whether the previous parameters are reused as initialization for the next task (warm restart) or not (cold restart).
 
 Furthermore, we consider the history size as a controlled parameter and vary it according to the percentiles of $\Dt{k}$, as determined in our analyses of the datasets in \Secref{sub:new-datasets}. 
 Corresponding to two layers of graph convolution, which our models use, the quartiles consider $25\%$, $50\%$, and $75\%$ of the $\Dt{2}$ distribution and are in terms of history sizes $c=1$, $3$, and $6$ for both DBLP datasets, and $1$, $4$, and $8$ for the PharmaBio dataset. 
 We compare these limited-history settings with full-graph training, which corresponds to keeping an unlimited history of the entire timeline of the graph.
 
All methods are trained in a transductive fashion, except for GraphSAINT, which needed to use the inductive setting. However, we have ensured that the evaluation is fair (see \Secref{sec:datasets}) and we have confirmed in Experiment~1 (see \Secref{exp:transductive-inductive}) that the difference between inductive and transductive training is negligible.

\subsection{Hyperparameters}
We constrain all models to two graph convolutional layers, a comparable penultimate hidden dimension (2x32 GraphSAGE, 4x8 GAT, 2x2x16 JKNet, 64 MLP), and a dropout rate of $0.5$. 
We fix an update step budget of $200$ per task and use Adam~\cite{Adam} to optimize cross-entropy. We implemented GAT, GraphSAGE-mean, SGC, and JKNet with \textit{dgl}~\cite{dgl} and use \textit{torch-geometric}~\cite{geometric} for GraphSAINT.
We had to disable GraphSAINT's norm recomputation for each task so that our experiments could finish in a reasonable time.

For each combination of a GNN, history size, and restart configuration, we tune the learning rate on \dblpeasy.
Thus, we consider \dblpeasy as our development dataset to tune the learning rate, which we then apply to \dblphard and \wos.
The search space for the learning rate is $\{0.1,0.05,0.01,0.005,$ $0.001,0.0005\}$.
We also optimized the weight decay, whose effect was negligible.

For the sake of a fair comparison, we have optimized the hyperparameters separately for each possible history size and restart configuration.

\subsection{Measures}
Our primary evaluation measure for lifelong vertex classification $f$ is accuracy.
With $\operatorname{acc}_t(f^\embrace{t})$, we denote the accuracy of model $f^\embrace{t}$ on task $\gT_t$. 
We aggregate the accuracy scores over the sequence of tasks $\gT_1, \ldots, \gT_T$ by using their unweighted average~\cite{DBLP:conf/nips/Lopez-PazR17}:
\begin{equation*}
    \operatorname{acc}(f) = \frac{1}{T} \sum_{t \in 1, \ldots, T} \operatorname{acc}_t(f^\embrace{t})
\end{equation*}
Following Lopez-Paz and Ranzato~\cite{DBLP:conf/nips/Lopez-PazR17}, we use Forward Transfer (FWT) to quantify the effect of reusing previous parameters.
This is reflected in the accumulated differences in accuracy between the $f_\text{warm}$ and $f_\text{cold}$ models, defined below:
\begin{equation*}
  \operatorname{FWT}(f_\text{warm}, f_\text{cold}) = \frac{1}{T-1} \sum_{t \in 2, \ldots,  T} \operatorname{acc}_t(f_\text{warm}^\embrace{t}) - \operatorname{acc}_t(f_\text{cold}^\embrace{t})
\end{equation*}

Experiments are repeated 10 times with different random seeds. We report the mean accuracy plus/minus 1.96 times the standard error of the mean.

\subsection{Results}
Table~\ref{tab:results} shows the aggregated results of 20,160 evaluation steps (48 configurations with 10 repetitions on two datasets with 12 tasks each and one dataset with 18 tasks). 
We consider the method $A$ to be better than the method $B$ when the mean accuracy of $A$ is higher than that of $B$ and the 95\% confidence intervals do not overlap~\cite{deeplearningbook}.
In terms of the absolute best methods per setting (= dataset $\times$ history size), we find that GraphSAGE consistently gives the highest scores except for \dblphard, where it is challenged by SGC. 

Regarding the comparison of history sizes (\ie \textit{explicit knowledge}, see \Secref{sec:introduction}), the highest scores are achieved in almost all cases by using an unlimited history size, \ie using the full graph's history.
However, in all datasets, the scores for training with limited window sizes larger than 1 are close to those for full-graph training.
With history sizes that cover 50\% of the GNN's receptive field, all methods achieve at least 95\% relative accuracy compared to the same model under full-history training.
When 75\% of the receptive field is covered, the models produce at least 99\% relative accuracy. To compute these percentages, we have selected the best of both cold and warm restarts for each method.

Regarding the influence of \emph{implicit knowledge}, we find that reusing parameters (warm restarts) leads to notably higher scores than retraining from scratch when the history size is one (see column FWT with history size $c=1$).
The average Forward Transfer across all models and datasets with history size $c=1$ is five accuracy points.

Regarding isotropic vs. anisotropic GNNs, we find that GAT and GraphSAGE perform similarly well on \dblpeasy (on which the learning rate was tuned). However, GraphSAGE-mean yields higher scores on \dblphard and \wos, which could indicate that GraphSAGE-mean is more robust to hyperparameters than GAT.

Regarding memory-efficient methods, we observe that the scores of SGC are among the highest of all methods on \dblphard. 
To understand this result, we recall that SGC uses only one single weight matrix of shape $n_\text{features} \times n_\text{outputs}$, which leads to 300,000 learnable parameters on \dblphard, but only 27,000 and 34,000 on \dblpeasy and \wos, respectively. 
SGC maps input features directly to classes, which results in a very high number of parameters on DBLP-Hard because this dataset has a high number of classes.
For comparison, GraphSAGE has $146,000$ learnable parameters on \dblpeasy, $264,000$ on \dblphard, and $310,000$ on \wos.
On the other hand, GraphSAINT yields scores on \wos comparable to GraphSAGE, but lower scores on both DBLP datasets.

\begin{sidewaystable}
\begin{center}
\begin{minipage}{\textheight}
\caption{Accuracy (with 95\% confidence intervals through 1.96 standard error of the mean) and Forward Transfer (averaged difference of warm and cold restarts) in our datasets with different history sizes (column \textbf{c}). The best method per case (= per one dataset and one history size) is marked in bold, along with the methods where the 95\% CI overlaps.}  
\label{tab:results}
\scriptsize
\begin{tabular*}{\textheight}{llccccccccc}
\toprule
& {} & \multicolumn{3}{c}{GAT} & \multicolumn{3}{c}{GraphSAGE-Mean} & \multicolumn{3}{c}{MLP (Baseline)} \\
          & {} &     \multicolumn{2}{c}{avg. accuracy} & FWT &     \multicolumn{2}{c}{avg. accuracy} & FWT &   \multicolumn{2}{c}{avg. accuracy} & FWT \\
          & {} &    cold & warm  & {} &     cold & warm & {}  &   cold & warm & {} \\

\textbf{Dataset} & \textbf{c} &               &          &               &          &               &          \\
\midrule
\multirow{4}{*}{\textbf{\dblpeasy{}}}
        & \textbf{1} & $60.8 \pm 0.5$ & $\mathbf{64.9 \pm 0.4}$ & $+4.5$ & $60.4 \pm 0.5$ & $\mathbf{65.1 \pm 0.4}$ & $+5.2$ & $56.1 \pm 0.4$ & $62.2 \pm 0.5$ & $+6.6$\\
        & \textbf{3} & $\mathbf{68.9 \pm 0.3}$ & $\mathbf{69.3 \pm 0.3}$ & $+0.2$ & $\mathbf{68.7 \pm 0.3}$ & $\mathbf{69.3 \pm 0.3}$ & $+0.7$ & $61.0 \pm 0.5$ & $62.9 \pm 0.4$ & $+2.0$\\
        & \textbf{6} & $\mathbf{70.3 \pm 0.4}$ & $70.2 \pm 0.4$ & $-0.1$ & $\mathbf{71.1 \pm 0.4}$ & $\mathbf{70.9 \pm 0.4}$ & $-0.3$ & $62.7 \pm 0.3$ & $62.7 \pm 0.4$ & $-0.2$\\
     & \textbf{full} & $70.2 \pm 0.4$ & $70.2 \pm 0.4$ & $+0.1$ & $\mathbf{71.6 \pm 0.4}$ & $\mathbf{71.4 \pm 0.3}$ & $-0.2$ & $63.4 \pm 0.3$ & $61.9 \pm 0.4$ & $-1.2$\\
          \midrule
\multirow{4}{*}{\textbf{\dblphard{}}} 
        & \textbf{1} & $39.4 \pm 0.2$ & $39.1 \pm 0.2$ & $-0.1$ & $34.5 \pm 0.4$ & $40.0 \pm 0.2$ & $+5.9$ & $31.6 \pm 0.3$ & $38.3 \pm 0.3$ & $+7.4$\\
        & \textbf{3} & $44.0 \pm 0.2$ & $43.7 \pm 0.2$ & $-0.4$ & $44.3 \pm 0.2$ & $\mathbf{45.1 \pm 0.2}$ & $+0.8$ & $33.7 \pm 0.3$ & $38.9 \pm 0.2$ & $+5.6$\\
        & \textbf{6} & $45.1 \pm 0.3$ & $45.3 \pm 0.3$ & $+0.2$ & $\mathbf{46.5 \pm 0.3}$ & $\mathbf{46.7 \pm 0.3}$ & $+0.2$ & $39.2 \pm 0.2$ & $38.3 \pm 0.2$ & $-0.7$\\
     & \textbf{full} & $45.6 \pm 0.3$ & $45.6 \pm 0.3$ & $-0.1$ & $46.8 \pm 0.2$ & $47.1 \pm 0.3$ & $+0.4$ & $38.2 \pm 0.2$ & $36.7 \pm 0.2$ & $-1.1$\\
          \midrule
\multirow{4}{*}{\textbf{\wos{}}} 
        & \textbf{1} & $61.6 \pm 0.9$ & $65.4 \pm 0.9$ & $+3.8$ & $65.4 \pm 0.9$ & $\mathbf{68.6 \pm 1.0}$ & $+3.3$ & $62.7 \pm 0.9$ & $66.3 \pm 0.9$ & $+3.9$\\
        & \textbf{4} & $64.5 \pm 0.8$ & $65.3 \pm 0.9$ & $+0.9$ & $\mathbf{68.0 \pm 0.8}$ & $\mathbf{69.0 \pm 0.8}$ & $+1.1$ & $66.3 \pm 0.7$ & $65.7 \pm 0.8$ & $-0.7$\\
        & \textbf{8} & $65.1 \pm 0.8$ & $65.4 \pm 0.8$ & $+0.3$ & $\mathbf{68.8 \pm 0.7}$ & $\mathbf{69.0 \pm 0.8}$ & $+0.2$ & $64.2 \pm 0.8$ & $65.3 \pm 0.7$ & $+0.9$\\
     & \textbf{full} & $64.3 \pm 0.8$ & $65.4 \pm 0.8$ & $+0.2$ & $\mathbf{69.0 \pm 0.7}$ & $\mathbf{68.4 \pm 0.7}$ & $-0.7$ & $65.4 \pm 0.8$ & $64.4 \pm 0.6$ & $-1.1$\\
          \midrule\\
          & {} & \multicolumn{3}{c}{SGC} & \multicolumn{3}{c}{GraphSAINT} & \multicolumn{3}{c}{Jumping Knowledge} \\
          & {} &     \multicolumn{2}{c}{avg. accuracy} & FWT &     \multicolumn{2}{c}{avg. accuracy} & FWT &   \multicolumn{2}{c}{avg. accuracy} & FWT \\
          & {} &    cold & warm  & {} &     cold & warm & {}  &   cold & warm & {} \\
\midrule
\multirow{4}{*}{\textbf{\dblpeasy}} 
        & \textbf{1} & $57.1 \pm 0.4$ & $63.7 \pm 0.4$ & $+7.2$ & $62.1 \pm 0.3$ & $63.2 \pm 0.4$ & $+1.2$ & $56.2 \pm 0.5$ & $61.4 \pm 0.5$ & $+5.6$\\
        & \textbf{3} & $66.4 \pm 0.3$ & $67.4 \pm 0.3$ & $+1.2$ & $66.4 \pm 0.4$ & $65.3 \pm 0.5$ & $-0.9$ & $65.2 \pm 0.3$ & $65.9 \pm 0.5$ & $+1.0$\\
        & \textbf{6} & $69.3 \pm 0.4$ & $69.3 \pm 0.4$ & $+0.1$ & $68.1 \pm 0.4$ & $65.5 \pm 0.7$ & $-2.1$ & $68.0 \pm 0.4$ & $66.9 \pm 0.6$ & $-0.7$\\
     & \textbf{full} & $\mathbf{71.0 \pm 0.4}$ & $70.0 \pm 0.4$ & $-1.0$ & $68.4 \pm 0.5$ & $65.7 \pm 0.5$ & $-2.8$ & $68.7 \pm 0.4$ & $66.3 \pm 0.4$ & $-2.5$\\
          \midrule
\multirow{4}{*}{\textbf{\dblphard}} 
        & \textbf{1} & $34.5 \pm 0.3$ & $\mathbf{41.0 \pm 0.3}$ & $+7.0$ & $35.9 \pm 0.3$ & $35.6 \pm 0.4$ & $+0.5$ & $33.0 \pm 0.2$ & $35.3 \pm 0.3$ & $+2.9$\\
        & \textbf{3} & $44.1 \pm 0.2$ & $\mathbf{44.8 \pm 0.3}$ & $+0.8$ & $39.3 \pm 0.3$ & $38.1 \pm 0.5$ & $-0.6$ & $39.1 \pm 0.3$ & $38.8 \pm 0.4$ & $+0.3$\\
        & \textbf{6} & $\mathbf{46.9 \pm 0.3}$ & $46.2 \pm 0.3$ & $-0.4$ & $40.6 \pm 0.3$ & $38.8 \pm 0.6$ & $-1.2$ & $41.0 \pm 0.3$ & $40.1 \pm 0.5$ & $-0.3$\\
     & \textbf{full} & $\mathbf{48.8 \pm 0.4}$ & $47.5 \pm 0.3$ & $-1.2$ & $41.0 \pm 0.4$ & $40.7 \pm 0.4$ & $-0.3$ & $41.6 \pm 0.3$ & $40.8 \pm 0.2$ & $-0.9$\\
          \midrule
\multirow{4}{*}{\textbf{\wos}} 
        & \textbf{1} & $62.3 \pm 0.9$ & $64.5 \pm 0.8$ & $+2.3$ & $65.7 \pm 0.8$ & $\mathbf{68.6 \pm 0.8}$ & $+3.0$ & $64.1 \pm 0.9$ & $\mathbf{68.3 \pm 0.9}$ & $+4.3$\\
        & \textbf{4} & $64.4 \pm 0.8$ & $64.4 \pm 0.8$ & $-0.0$ & $67.3 \pm 0.8$ & $\mathbf{68.4 \pm 0.7}$ & $+1.0$ & $67.1 \pm 0.8$ & $\mathbf{68.2 \pm 0.8}$ & $+1.1$\\
        & \textbf{8} & $65.3 \pm 0.8$ & $64.0 \pm 0.7$ & $-1.4$ & $\mathbf{68.1 \pm 0.8}$ & $\mathbf{68.0 \pm 0.7}$ & $-0.1$ & $\mathbf{67.8 \pm 0.8}$ & $\mathbf{67.7 \pm 0.7}$ & $-0.3$\\
     & \textbf{full} & $62.4 \pm 0.8$ & $61.7 \pm 0.6$ & $-0.8$ & $\mathbf{68.2 \pm 0.8}$ & $66.1 \pm 0.8$ & $-2.2$ & $66.8 \pm 0.8$ & $64.5 \pm 0.7$ & $-2.6$\\
\bottomrule
\end{tabular*}
\end{minipage}
\end{center}
\end{sidewaystable}

\subsection{Ablation Study: Incrementally-Trained vs. Once-Trained Models}\label{sub:ablation}
In contrast to retraining with different history sizes, one may also wonder how long a once-trained model generalizes. 
Thus, we have analyzed how long a model, which is trained only once at a specific point in time, will generalize well over a sequence of tasks over time.
We isolate the effect of incremental training and compare once-trained trained models (static) with incrementally trained models (incremental).
Static models are trained for 400 epochs on the data before the first evaluation time step, which comprises 25\% of the total \nodes.
Incrementally trained models are trained for 200 epochs with history sizes of 3 timesteps (4 on the \wos dataset) before evaluating each task. We repeat each experiment 10 times with different random seeds.
\begin{figure*}[ht!]
\centering
  \includegraphics[width=1\linewidth]{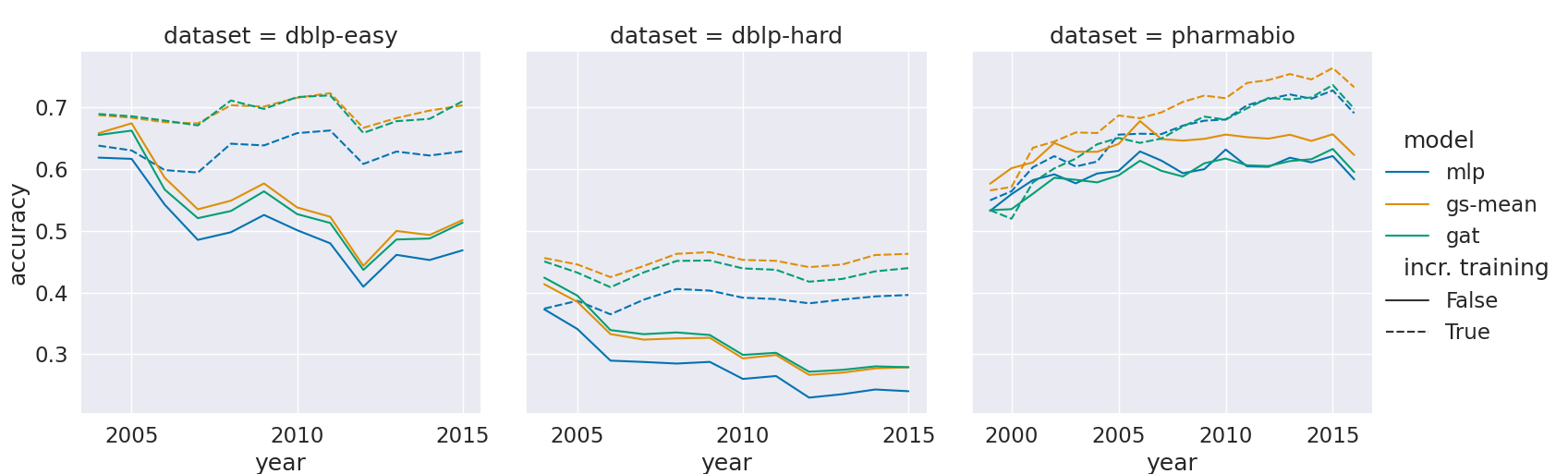}
  \caption{Results of the ablation study: Accuracy scores of once-trained, static models (solid lines) are lower than incrementally trained models (dashed lines). 
  }\label{fig:exp0}
\end{figure*}
In \Figref{fig:exp0}, we see that the accuracy of the static
models decreases with time on \dblpeasy and \dblphard, where new classes appear
over time. On \wos (fixed class set), the accuracy of the static models plateaus,
whereas the accuracy of incrementally trained models increases. 
We see that incremental training is not only necessary to adapt to new classes, but also helpful to make use of an increased amount of training data.

\subsection{Summary}
This experiment shows that in the three analyzed datasets, with only history sizes of 3 or 4 (corresponding to 50\% coverage of the receptive field of a 2-layer GCN), almost all methods obtain 95\% accuracy compared to the same model under full-history training.
Moreover, with very small history sizes, such as using only one past task, using warm restarts is important to maintain a high level of accuracy. Furthermore, we have confirmed in an ablation study that incremental training is necessary to account for changes of the graph.

\section{Experiment 3: Lifelong Learning with Limited Labeled Data}\label{exp:limited-labeled-data}

Until now, we have assumed that the true labels of \nodes become part of the training data for subsequent tasks. Now we relax this assumption and release only a portion of the labeled data in task $t$ for training in the subsequent task $t+1$. 
This resembles real-world applications, such as the indexing of scientific articles in libraries~\cite{DBLP:conf/jcdl/MaiGS18}. 
The motivation is that labeled data is expensive to ``produce''.
Again, we work with the most challenging dataset, \dblphard{}, for this experiment, because it has the highest number of new classes.

\subsection{Procedure}
To implement the idea of learning with only a fraction of labeled data, we randomly sample a subset of \nodes, for which true class labels are available for training. We denote this fraction as \emph{label rate}.
For the experiments, it is important to sample globally rather than on a per-task basis to avoid nodes toggling between being labeled and unlabeled.
Therefore, we sample the entire dataset before splitting it into tasks. In this way, the subset of \nodes that comes with classes is fixed for the entire duration of the experiments. Furthermore, we used the same subset of classes with all different configurations and all repetitions of the experiment. We sample uniformly at random on the \node level without any stratification between classes. \revise{}{Note that this problem statement of testing different label rates is similar to the difference between settings A and B in Experiment 1 of \Secref{exp:transductive-inductive}. However, here we test the influence of the label rate in the context of a task sequence (instead of comparing only two tasks) and systematically change the label rates ranging from $0.1$ to $0.9$ (instead of only one ``split'').}

For this experiment, we use GraphSAGE-Mean as the GNN model because it achieved the best results in the previous experiment, where the label rate was not restricted. 
As above, we experiment with different history sizes 1, 3, and 6 and both restart configurations warm and cold. As the dataset, we use \dblphard because it has the highest number of classes both in total and new classes that appear over time, and thus is the most challenging.

\subsection{Hyperparameters}
Again, as in the previous experiment, the optimal hyperparameters were determined on \dblpeasy{}, the sub-dataset of \dblphard that we consistently use to tune hyperparameters.
The search space for the learning rate is again $\{0.1,0.05,0.01,0.005,0.001,0.0005\}$. We have not tuned the hyperparameters separately for each label rate, but we reuse the optimal hyperparameters from training with a 100\% label rate.

\subsection{Measures}
As in the previous experiment, we use the average accuracy across tasks as the evaluation metric.

\subsection{Results}
\begin{figure}
    \centering
    \includegraphics[width=0.97\linewidth]{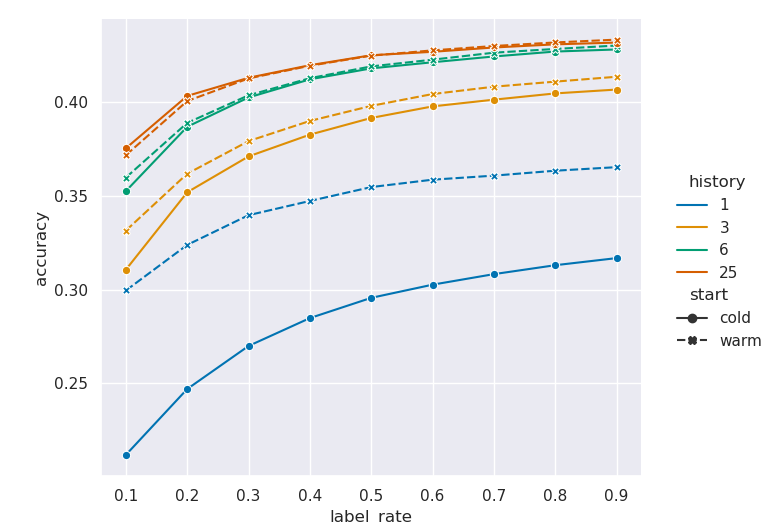}
    \caption{Average accuracy of GraphSAGE with warm restarts across tasks on \dblphard under varying label rate}
    \label{fig:label_rate}
\end{figure}

In \Figref{fig:label_rate}, we plot the average accuracy between tasks as a function of the label rate. 
As expected, the absolute accuracy values decrease as the label rate decreases.
However, we made a similar observation as in previous experiments with respect to warm/cold restarts. Using warm restarts consistently leads to higher scores than cold restarts. The effect is more prolonged when the history size is small.

When comparing history sizes, we again observe that a larger history size leads to better results. In particular, using the entire history gives the best results, closely followed by a history size of 6. Still, when the label rate is decreased, the difference between the history sizes remains constant.

With very low label rates (in the range between 10\% and 30\%), the accuracy of the cold restart strategy drops faster than the accuracy of warm restarts. In other words, the use of warm restarts leads to more stable models when dealing with lower label rates.

\subsection{Summary}
This experiment shows that the effect of varying the label rate is as expected: the performance degrades with fewer labeled training data. We confirm the finding from previous experiments that warm restarts consistently lead to higher performance than cold restarts when the history size is small. 
Furthermore, we observe that warm restarts become even more important when the label rate is low. 

\section{Experiment 4: Detection of Unseen Classes}\label{exp:open-learning}
In our evolving graphs, we have to deal with previously unseen classes.
In previous experiments on lifelong learning, these unseen classes (and 
the vertices that have these classes) were already part of the test data. 
However, the models did not have the opportunity to actually predict those classes, as no dedicated technique has been used to detect \nodes from unseen classes. 
Here, we evaluate our adaptation of the unseen class detection method DOC to graph data, called gDOC, as introduced in \Secref{sub:methods:unseen}. The experiments comprise a crisp unsupervised detection of instances of unseen classes. At the same time, the models need to make predictions as usual for the known classes.

\begin{figure}
    \centering
    \includegraphics[width=0.9\columnwidth]{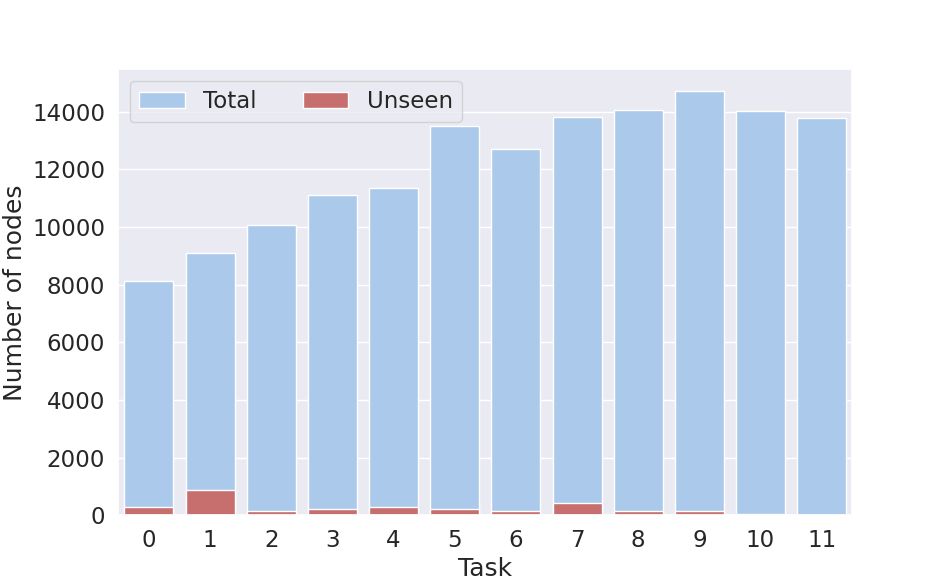}
    \caption{Number of \nodes with unseen classes per task on \dblphard}
    \label{fig:unseen_per_task}
\end{figure}

\subsection{Procedure}
In previous experiments, unseen classes were part of the test data, while there was no active treatment of having them detected automatically.
In this experiment, we seek to evaluate the performance of the gDOC method to detect unseen classes. As before, we train on task $t-1$ and evaluate on $t$ over a sequence of $T$ tasks.
However, for each \node, we use our unseen class detection module gDOC to predict whether this \node belongs to a previously known class or not.
If the prediction is that the \node does not belong to any previously known class, we reject its classification and assign a special virtual class (``unseen'').
As unseen class detection modules, we compare the original DOC as a baseline with our proposed gDOC method.

We used the \dblphard dataset, which has 23 new classes.
In addition to the dataset analysis in \Secref{sub:new-datasets}, we show in  \Figref{fig:unseen_per_task} how many \nodes belong to unseen classes in the \dblphard dataset. We also experiment with DBLP-easy, which has 4 new classes.
We use the best-performing model GraphSAGE-mean along with gDOC for unseen class detection that we have introduced in \Secref{sub:methods:unseen}. Our baseline is the original DOC method, also applied to the outputs of GraphSAGE-mean.
We observe that in every task except for the last one, there are \nodes with unseen classes.

\subsection{Hyperparameters}
As in the previous experiments, we optimize the model hyperparameters in our development data set \dblpeasy{}. We repeat the hyperparameter optimization because the loss function has changed from categorical to binary cross-entropy.
As before, the best learning rate is selected based on the best accuracy on \dblpeasy and transferred to \dblphard.

Note that we did not tune the learning rate for unseen class detection performance, but for the best accuracy, as in previous experiments. We then compare DOC with gDOC, where the former is our baseline and the latter uses our proposed class weighting loss function for lifelong learning.

\subsection{Measures}
We evaluate how well the models detect unseen classes. For this purpose, we use two measures: Macro-F1 with a special class for instances of unseen classes~\cite{DOC} and the Matthews correlation coefficient (MCC).
Note that Macro-F1 averages the F1 scores over classes such that the effect of the 'unseen' class is taken into account as any of the known classes.
In detail, we compute this \emph{Open Macro-F1} as 

\begin{align*}
    \text{Open Macro-F1} &:= \frac{1}{T} \sum_{t=1}^{T}\text{Macro-F1}(\vy^\prime{(t)}, \vy_{\text{pred}}^\prime{(t)})
\end{align*}

with

\begin{align*}
    \vy_{\text{pred},i}^\prime &:=
    \begin{cases}
     \text{'unseen'} \text{, if example }  i \text{ is detected as OOD}\\
    \vy_{\text{pred},i}, \text{ otherwise}\\
    \end{cases}\\
    \vy_i^\prime &:=
    \begin{cases}
    \vy_i \text{ if class } \vy_i \text{ is known}\\
    \text{'unseen'}, \text{ otherwise}\\
    \end{cases}\\
\end{align*}

where $\vy_i$ are the true labels and $\vy_\text{pred}$ are the predicted class labels.
The $\arg \max$ of the output is replaced by a special symbol when the method has emitted an 'unseen' decision for that instance.
The true labels $\vy$ are preprocessed similarly so that instances of previously unseen classes receive a special class symbol.

\begin{figure}
    \centering
    \includegraphics[width=0.7\linewidth]{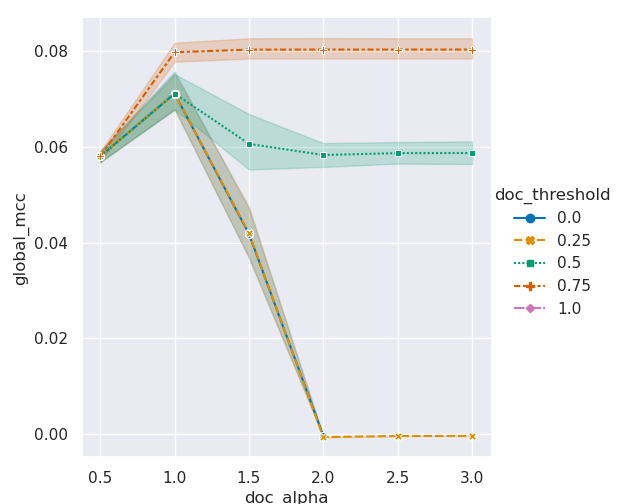}
    \caption{MCC score of gDOC with GraphSAGE-mean as GNN model (history size 3, warm restart setting) as a function of the risk reduction factor $\alpha$ and varying \emph{minimum} threshold values. We observe that the more risk reduction does not improve the results.}
    \label{fig:alpha_ablation}
\end{figure}

In pre-experiments, we found that the best Open F1-Macro scores are achieved when the thresholds are high. This is because we have a high number of classes and the special class contributes only very little to the overall F1 Macro score. 
Thus, a large number of false rejects, \ie a reject despite the class being known, diminishes the overall performance in terms of F1-Macro.

The F1-Macro score is limited in its expressiveness with respect to the detection of unseen classes since there are only a few vertices of that unseen class.
Thus, we report a further score, the Matthews correlation coefficient (MCC) of the 'unseen' class vs. all other classes (\ie the set of known classes).
MCC is a popular measure for evaluating binary classification that accounts for the class imbalance~\cite{chicco2020advantages}.
Dealing with this class imbalance is important as the number of vertices from the known classes is much larger than the number of vertices from the unseen class.
It ranges from -1 to 1, where zero corresponds to a random prediction.
In more detail, the MCC is computed as:
\begin{equation*}
    \mathrm{MCC} = \frac{\mathrm{TP} \cdot \mathrm{TN} - \mathrm{FP} \cdot \mathrm{FN}}{\sqrt{(\mathrm{TP} + \mathrm{FP})(\mathrm{TP}+\mathrm{FN})(\mathrm{TN}+\mathrm{FP})(\mathrm{TN}+\mathrm{FN})}}
\end{equation*}
where $\mathrm{TP}$ are true positives or correctly rejected instances, $\mathrm{TN}$ true negatives, $\mathrm{FP}$ false positives, and $\mathrm{FN}$ false negatives. We accumulate those numbers over the entire sequence of tasks.

\begin{table}[htp]
    \centering
    \caption{Results for unseen class detection on \textbf{\dblpeasy} with GraphSAGE as base model (average of 5 repetitions).  $\alpha$ indicates that risk reduction is used with the respective factor for the standard deviation, $\tau$ is the minimum threshold.
    Runs named gDOC are trained with weighted cross entropy. DOC is our baseline.}\label{tab:results:open:easy}
    \small
    \begin{tabular}{llcccc}
    \toprule
     & & \multicolumn{2}{c}{\textbf{MCC}} & \multicolumn{2}{c}{\textbf{Open F1 Macro}}\\
     & & cold & warm & cold & warm\\
     \textbf{c} & \textbf{Open Learning Method} & &  & \\
    \midrule
        1 & DOC ($\tau=0.50$)            &.05 & .07 & .25 & .25\\
          & DOC ($\tau=0.50,\alpha=3.0$) & .05 & .07 & .25 & .25\\
          & gDOC ($\tau=0.50$)            & .04 & \textbf{.08} & .30& \textbf{.33}\\
          & gDOC ($\tau=0.50,\alpha=3.0$) & .04 & .05 & .30 & .32\\
          & gDOC ($\tau=0.75$)            & .04 & .07 & .30 & .30\\
        \midrule
        3 & DOC ($\tau=0.50$)            &.05 & .05 & .28 &.30\\
          & DOC ($\tau=0.50,\alpha=3.0$) & .05 & .05 & .28 & .30\\
          & gDOC ($\tau=0.50$)            & .05 & .08 & .34 & .34\\
          & gDOC ($\tau=0.50,\alpha=3.0$) & .06 & .08 & .34 & .34\\
          & gDOC ($\tau=0.75$)            & .07 & \textbf{.09} & .34 & .34\\
        \midrule
        6 & DOC ($\tau=0.50$)            & .06 & .06 &.31 & .32\\
          & DOC ($\tau=0.50,\alpha=3.0$) & .06 & .06 & .31 & .32\\
          & gDOC ($\tau=0.50$)            & .07 & .07 & .35 & .35\\
          & gDOC ($\tau=0.50,\alpha=3.0$) & .07 & .07 & .35 & .35\\
          & gDOC ($\tau=0.75$)            & .09 & \textbf{.10} & .35 & .35\\
        \midrule
        full & DOC ($\tau=0.50$)     & .07 & .07 & .32 & .33\\
          & DOC ($\tau=0.50,\alpha=3.0$) & .07 & .07 & .32 & .33\\
          & gDOC ($\tau=0.50$)            & .06  & .06 & .35 & .35\\
          & gDOC ($\tau=0.50,\alpha=3.0$) & .06  & .06 & .35 & .35\\
          & gDOC ($\tau=0.75$)            & .08   & \textbf{.10} & .35 & .35\\
        \bottomrule\\
    \end{tabular}
\end{table}

\begin{table}[htp]
    \centering
    \caption{Results for unseen class detection on \textbf{\dblphard} with GraphSAGE as base model (average of 5 repetitions).  $\alpha$ indicates that risk reduction is used with the respective factor for the standard deviation, $\tau$ is the minimum threshold.
    Runs named gDOC are trained with weighted cross entropy. DOC is our baseline.}\label{tab:results:open}
    \begin{tabular}{llcccc}
    \toprule
     & & \multicolumn{2}{c}{\textbf{MCC}} & \multicolumn{2}{c}{\textbf{Open F1 Macro}}\\
     & & cold & warm & cold & warm\\
     \textbf{c} & \textbf{Open Learning Method} & &  & \\
    \midrule
        1 & DOC ($\tau=0.50$)            & .01 & .04 & .01 & .01\\
          & DOC ($\tau=0.50,\alpha=3.0$) & .01 & .02 & .01 & .01\\
          & gDOC ($\tau=0.50$)            & .04 & .05 & \textbf{.13} & \textbf{.13}\\
          & gDOC ($\tau=0.50,\alpha=3.0$) & .04 & .05 & \textbf{.13} & \textbf{.13}\\
          & gDOC ($\tau=0.75$)            & .04 & \textbf{.09} & \textbf{.13} & \textbf{.13}\\
        \midrule
        3 & DOC ($\tau=0.50$)            & .02 & .03 & .02 & .05\\
          & DOC ($\tau=0.50,\alpha=3.0$) & .02 & .03 & .02 & .05\\
          & gDOC ($\tau=0.50$)            & .05 & .06 & \textbf{.15} & \textbf{.15}\\
          & gDOC ($\tau=0.50,\alpha=3.0$) & .05 & .06 & \textbf{.15} & \textbf{.15}\\
          & gDOC ($\tau=0.75$)            & .05 & \textbf{.08} & \textbf{.15} & \textbf{.15}\\
        \midrule
        6 & DOC ($\tau=0.50$)            & .02 & .03 & .05 & .08\\
          & DOC ($\tau=0.50,\alpha=3.0$) & .02 & .03 & .05 & .08\\
          & gDOC ($\tau=0.50$)            & .05 & .06 & \textbf{.16} & \textbf{.16}\\
          & gDOC ($\tau=0.50,\alpha=3.0$) & .05 & .06 & \textbf{.16} & \textbf{.16}\\
          & gDOC ($\tau=0.75$)            & .05 & \textbf{.07} & \textbf{.16} & \textbf{.16}\\
        \midrule
        full & DOC ($\tau=0.50$)     & .02 & .04 & .08 & .12\\
          & DOC ($\tau=0.50,\alpha=3.0$) & .02 & .04 & .08 & .12\\
          & gDOC ($\tau=0.50$)            & .04 & .05 & \textbf{.16} & \textbf{.16}\\
          & gDOC ($\tau=0.50,\alpha=3.0$) & .05 & .05 & \textbf{.16} & \textbf{.16}\\
          & gDOC ($\tau=0.75$)            & .05 & \textbf{.07} & \textbf{.16} & \textbf{.16}\\
        \bottomrule\\
    \end{tabular}
\end{table}

\subsection{Results}
The results for \dblpeasy{} are shown in Table~\ref{tab:results:open:easy} and the results for \dblphard{} in Table~\ref{tab:results:open}.
For both \dblpeasy{} and \dblphard{}, we observe that the MCC scores, which measure the correct detection of new classes, are consistently higher for gDOC than for plain DOC. The same holds for the Open F1 Macro scores, which measure the overall performance of OOD detection + classification: gDOC is consistently better than plain DOC.

When comparing \dblpeasy{} and \dblphard{}, we see that the absolute F1 score and the MCC score attained on \dblpeasy{} are higher than the absolute scores on \dblphard{}, which is expected since \dblphard{} has more classes and \dblpeasy{} is the subset of data on which hyperparameters were tuned. 

On \dblphard{}, the F1 score of plain DOC with a limited history size is very low: between 0.01 for history size 1 and 0.12 for unlimited history size. In the same setting, gDOC achieves much higher scores: already 0.13 with a history size of 1 and 0.16 with at least a history size of 6.
This shows that the class-weighted binary cross-entropy in gDOC is necessary to achieve reasonable F1 scores.

For thresholds, the results indicate that a high threshold (0.75) is preferable to lower thresholds.
We further note that the combination of warm restarts and a small history size leads to the highest MCC score (0.09) on \dblphard{}, while on \dblpeasy{}, on which the hyperparameters have been tuned, the MCC score is higher for larger history sizes.

In \Figref{fig:alpha_ablation}, we show that risk reduction, \ie lowering the detection threshold based on the class-specific standard deviation, does not help to increase performance. With a low minimum threshold (\eg 0), we see the pure performance of the risk reduction technique, which peaks at $\alpha=1$ before it decreases.
When using a high minimum threshold (0.5, 0.75, 1.0), applying risk reduction only decreases the OOD performance. In other words, the absolute best OOD detection performance is achieved when the minimum threshold $\tau$ is set to 0.75, regardless of the risk reduction factor $\alpha$. Therefore, the usefulness of risk reduction for our heavily imbalanced datasets is questionable.

To understand this result, we recall that risk reduction is a technique for calculating class-specific thresholds $\tau_i$ (see \Secref{sub:methods:unseen}).
However, this is only possible up to the global minimum threshold $\tau$.
Thus, even with risk reduction, class-specific thresholds cannot go below $\tau$.

\revise{}{\subsection{Combining gDOC with Different GNN Base Models}\label{sub:gdoc-base-model-comparison}}
\revise{}{
The gDOC module can be used in conjunction with arbitrary GNN base models.
We compare GraphSAGE, GAT, and SGC as a base model for gDOC. 
We chose SGC because of its strong performance on the DBLP-hard dataset in Experiment 2, along with GAT as the most popular anisotropic model and GraphSAGE since it is one of the most popular isotropic models. 
As in the previous experiments, the learning rate was tuned for ID classification on \dblpeasy{}.
Each configuration of history size and cold/warm restarts is independently optimized with respect to the hyperparameters.}

\revise{}{
The results are shown in Table~\ref{tab:gdoc-base-model-comparison}. The ranking of the base models is similar to the results obtained in Experiment 2, which shows that adding the gDOC module has no unexpected effects on the base models. In particular, using SGC leads to similar performance as GraphSAGE. However, SGC exceeds 30\,GB GPU memory on the full-history configuration and runs out of memory. GAT performs below GraphSAGE and SGC under smaller history size conditions, while it tends to catch up in terms of in-distribution accuracy and Macro-F1 when more history is available. In conclusion, this comparison confirms that gDOC can be successfully combined with various GNN base models.}

\begin{table}[ht]
    \centering
    \caption{Comparison of gDOC combined with different base models on \dblphard{}. The gDOC threshold is set to the 0.75 and no risk reduction is applied.}\label{tab:gdoc-base-model-comparison}
    \begin{tabular}{llcccccc}
    \toprule
     & & \multicolumn{2}{c}{\textbf{ID Accuracy}} & \multicolumn{2}{c}{\textbf{OOD MCC}} & \multicolumn{2}{c}{\textbf{Open F1}}\\ 
     & & cold & warm & cold & warm & cold & warm\\
     \textbf{c} & \textbf{Method} \\
    \midrule
        1 & GS+gDOC & \textbf{36.4} & 37.6 & .04 & .09 & \textbf{.13} & .13\\
        & SGC+gDOC & 35.0 & \textbf{38.4} & \textbf{.05} & \textbf{.10} & .12 & \textbf{.14}\\
        & GAT+gDOC & 34.0 & 38.0 & .04 & .08 & .10 & .13\\
        \midrule
        3 & GS+gDOC & 40.9 & 40.7 & \textbf{.05} & \textbf{.08} & \textbf{.15} & .15\\
        & SGC+gDOC & \textbf{41.6} & \textbf{41.9} & .05 & .07  & \textbf{.15} & \textbf{.16}\\
        & GAT+gDOC & 40.3 & 40.3 & .04 & .07 & .13 & .13\\
        \midrule
        6 & GS+gDOC & 42.5 & 42.2 & \textbf{.05} & \textbf{.07} & \textbf{.16} & \textbf{.16}\\
        & SGC+gDOC & \textbf{43.7} & \textbf{43.4} & .04 & \textbf{.07}  & \textbf{.16} & \textbf{.16}\\
        & GAT+gDOC & \textbf{43.7} & \textbf{43.4} & .04 & \textbf{.07} & \textbf{.16} & \textbf{.16}\\
        \midrule
        full & GS+gDOC & 43.6 & \textbf{43.5} & \textbf{.05} & \textbf{.07} & \textbf{.16} & \textbf{.16} \\
        & SGC+gDOC & \multicolumn{6}{c}{\textit{out of GPU memory}}\\
        & GAT+gDOC & \textbf{43.9} & \textbf{43.5} & .04 & .05 & \textbf{.16} & \textbf{.16}\\
       \bottomrule 
    \end{tabular}
\end{table}

\revise{}{\subsection{Trade-off between In-Distribution Accuracy and OOD Detection}\label{sub:gdoc-tradeoff}}
\revise{}{We assess how in-distribution accuracy is affected by new class detection capabilities. Therefore, we report the average accuracy across tasks, calculated in the same way as in Experiment 2 from \Secref{exp:lifelong-learning}.}
\revise{}{The results are reported in Table~\ref{tab:gdoc_ablation} and show that, as expected, a plain GraphSAGE without OOD capabilities has a higher in-distribution accuracy than training with OOD detection capabilities (GraphSAGE+gDOC). This difference is caused by training with binary cross-entropy instead of the standard categorical cross-entropy. An interesting exception is that GraphSAGE+gDOC is better than GraphSAGE on the smallest history size configuration ($c=1$). We assume that this difference is caused by GraphSAGE overfitting to the little data from a single graph snapshot, whereas the weighted cross-entropy of gDOC seems to alleviate this problem.}

\begin{table}[ht]
    \centering
    \caption{Trade-off between in-distribution classification accuracy and out-of-distribution detection performance on \dblphard{}. GraphSAGE (without an OOD detection module) is trained with categorical cross-entropy, while the methods capable of OOD detection are trained with binary cross-entropy. For ID accuracy, we always select the class with the maximum logit, regardless of any OOD threshold. NA marks no OOD detection capabilities.}\label{tab:gdoc_ablation}
    \begin{tabular}{llcccc}
    \toprule
     & & \multicolumn{2}{c}{\textbf{ID Accuracy}} & \multicolumn{2}{c}{\textbf{OOD MCC}}\\ 
     & & cold & warm & cold & warm\\
     \textbf{c} & \textbf{Method} \\
    \midrule
        1 & GraphSAGE+gDOC($\tau=0.75$) & \textbf{36.4} & 37.6 & \textbf{.04} & \textbf{.09}\\
        & GraphSAGE+DOC($\tau=0.5, \alpha=3.0)$ & 35.2 & 28.7 & .01 & .02\\
        & GraphSAGE & 34.5 & \textbf{40.0} & NA & NA\\
        \midrule
        3 & GraphSAGE+gDOC($\tau=0.75$) & 40.9 & 40.7 & \textbf{.05} & \textbf{.08}\\
        & GraphSAGE+DOC($\tau=0.5, \alpha=3.0$) & 39.4 & 43.1 &.02 & .03\\
        & GraphSAGE & \textbf{44.3} & \textbf{45.1} & NA & NA\\
        \midrule
        6 & GraphSAGE+gDOC($\tau=0.75$) & 42.5 & 42.2 & \textbf{.05} & \textbf{.07}\\
        & GraphSAGE+DOC($\tau=0.5, \alpha=3.0$) & 43.6 & 44.1 & .02 & .03\\
        & GraphSAGE & \textbf{46.5} & \textbf{46.7} & NA & NA\\
        \midrule
        full & GraphSAGE+gDOC($\tau=0.75$) & 43.6 & 43.5 & \textbf{.05} & \textbf{.07}\\
        & GraphSAGE+DOC($\tau=0.5, \alpha=3.0$)& 42.9 & 45.1 & .02 & .04\\
        &  GraphSAGE & \textbf{46.8} & \textbf{47.1} & NA & NA \\
        \bottomrule
        
    \end{tabular}
\end{table}

\subsection{Summary}
Our experiments have shown that weighting the binary
cross-entropy loss function in gDOC is essential for unseen class detection in
imbalanced graph data.  We also learned that the risk reduction technique (as
proposed in DOC~\cite{DOC}) is not helpful on our imbalanced graph datasets.
That is because the variance among predictions in the unbalanced case is so
high that the (minimum) threshold effectively never changed. The only
exceptions are tiny factors of standard deviation ($< 1$).  Nevertheless, this
only decreases the unseen class detection performance measured by MCC. We
recommend using gDOC with weighted binary cross-entropy to account for class
imbalance. However, we could not find any benefits of the risk reduction
technique proposed in the original DOC. \revise{}{We have successfully combined
gDOC with different base models (see \Secref{sub:gdoc-base-model-comparison})
and analyzed the trade-off between ID accuracy and OOD detection capabilities
(see \Secref{sub:gdoc-tradeoff}).}

\section{General Discussion}\label{sec:discussion}

\subsection{Main Findings}

Our experiments show several key results. First, we have shown in \Secref{exp:transductive-inductive}  that it is \emph{not} necessary to up-train GNNs when new unlabeled data arrives. Instead, the performance of inductively pre-trained GNNs remains stable, even when new unlabeled data are added to the graph.

From the incremental training experiments with limited history sizes in
\Secref{exp:lifelong-learning}, we obtain results that are almost as good as when using the entire history of the graph: 
With window sizes of 3 or 4 (50\% receptive field coverage), GNNs achieve at least 95\% accuracy compared to using all past data for incremental training. With window sizes of 6 or 8 (75\% receptive field coverage), the GNN retains at least 99\% accuracy. This result holds for standard GNN architectures and scalable and sampling-based approaches.
This result directly impacts 
lifelong learning of GNNs in evolving graphs, as the setting closely resembles real-world applications.
We have investigated whether to reuse parameters from previous tasks (warm restarts). 
We find that reusing an ``old'' model is a viable strategy, even though new classes appear during the sequence of tasks and the history size is limited.
We have shown that reusing parameters from previous tasks becomes critical when the history sizes are small because less explicit knowledge is available.

We have shown in \Secref{exp:limited-labeled-data} that the methods work well, even when the labeled data are limited, which is essential for real-world applications because data annotation is expensive.

With the introduction of gDOC, we have made the first step to introduce new class detection in lifelong graph learning in \Secref{exp:open-learning}, by combining graph neural networks with the DOC~\cite{DOC} module and extending it to take into account class imbalance.
Our experiments on new class detection show that it is necessary to adjust the weights of binary cross-entropy training in gDOC to account for the imbalanced label distribution.
Contrary to the original DOC, we have not observed any improvements with risk reduction through the standard deviation of logits.
Instead, the best results were achieved with an appropriate threshold ($\tau=0.75$) regardless of the risk reduction factor $\alpha$. 
We acknowledge that emitting a crisp decision in unsupervised unseen class detection is a highly challenging problem. 

Another interesting result is that combining warm restarts with small history sizes has increased MCC scores on the most challenging \dblphard{} dataset. 
It seems that omitting old data helps to detect out-of-distribution examples better.

\subsection{Generalizability}
We have shown that our incremental training approach can be applied to various GNN models and is orthogonal to sampling and preprocessing approaches. 
Our incremental training procedure can generally be applied to any GNN architecture with few caveats. 
If the GNN architecture depends on transductive learning, this constraint carries over to incremental training. 
Similarly, any pre-computation steps, such as computing normalizing constants such as in GCN~\cite{DBLP:journals/corr/KipfW16} or GraphSAINT~\cite{DBLP:conf/iclr/ZengZSKP20}, must be performed again when adapting the model to a new task.

\revise{}{
We assume in this work that old data do not change, \eg the vertices' labels remain the same over time. 
This is a reasonable assumption for citation and collaboration graphs.
However, changes to old data may happen when generalizing the framework to other domains.
This is addressed by the framework as follows:
We use a certain history of the data for training defined by the history size $c$. Any change on older vertices like a label being changed would be immediately reflected in the following training iterations, \ie the next tasks. 
In the cold start setting, the next trained model would be immediately be trained on the new correct ground truth data (up to history size). 
In the warm restart setting, the old parameters could still encode the impure knowledge.
But ultimately it would also receive the up-to-date ground truth label as the training data for the next tasks. 
}

To reflect our work in the broader context of lifelong or continual learning, we reconsider the
gradient episodic memory framework~\cite{DBLP:conf/nips/Lopez-PazR17} for image data, in which the examples are independent. 
Specific pre-processing steps are required to cast graph data into independent examples for \node classification, such as transforming each \node into a graph~\cite{wangLifelongGraphLearning2021}.
This increases the number of inference steps by $\mathcal{O}(\lvert V \rvert)$ compared to our approach. 

\section{Conclusion and Future Work}

We have conducted extensive experiments to investigate how graph neural networks behave in a lifelong learning setting on evolving graph data in which the class distribution is highly imbalanced and the models need to adapt to new classes over time.
In the first experiment, we have shown that it is not necessary to up-train GNNs on new unlabeled data. Based on this result, we have explored in a second experiment the case of an evolving graph in which new labeled vertices are continually added, including new
classes over a sequence of tasks. 
The results show that parameter reuse allows us to retain a high level of accuracy, even with a limited history size.
In the third experiment, we continued in this setup and tested the sensitivity to the label rate in the evolving graph setup, where we confirmed our previous finding.
Lastly, in the fourth experiment, we compared our newly proposed gDOC extension against the simple adaption of DOC to graphs, showing that taking the class imbalance into account during training is crucial.
\revise{}{We have shown that gDOC can be successfully combined with different GNN models.}
To facilitate our analyses, we have shown that the $\Dt{k}$ measure to derive the history sizes is equivariant to different temporal granularities.
\revise{}{The measure $\Dt{k}$ quantifies the temporal differences along the edges in a temporal graph and is suitable to be reused independently from the other methods presented in this work.}
These results show a rich picture covering numerous challenges of applying graph neural networks in practical settings without retraining the model from scratch as soon as new data arrive.

As future work, we intend to explore and adapt more out-of-distribution approaches to graphs, \eg by using the IsoMax loss function~\cite{macedo2021improving}. \revise{}{Another promising direction of future work would adapt ideas from the L2AC framework to graphs, \ie integrating explicit retrieval and similarity components.} For the scope of this work, we have limited ourselves to techniques that provide a crisp decision rather than an OOD score because an OOD score requires validation data to tune the thresholds.  Instead, the crisp unseen class detection methods presented here will apply directly to real-world applications. 
Next, it will be interesting to analyze why omitting old training data helps detect out-of-distribution examples. 
Although we have removed old data solely based on the \node's time, future work might want to analyze different approaches to determine which \nodes to keep and which to remove, given a limited ``memory'' budget. For example, keeping \nodes with a high degree or page rank could be beneficial.
Another direction of future work would be to explore when it is safe to actively shrink the output layer of the GNNs, \eg by looking at the final layer's weights.
We envision that the results of this work will spur the development of new specialized techniques for lifelong open-world learning in evolving graphs.

\section*{Availability of Data and Code}

\begin{itemize}
\item Availability of data: We published our lifelong graph learning datasets at \href{https://doi.org/10.5281/zenodo.3764770}{https://doi.org/10.5281/zenodo.3764770}.
\item Availability of code: An implementation of our experimental framework is available at \href{https://github.com/lgalke/lifelong-learning}{https://github.com/lgalke/lifelong-learning}

\end{itemize}

\noindent

\bibliography{references}

\end{document}